\pdfoutput=1
\documentclass[11pt]{article}
\usepackage[table]{xcolor}
\usepackage{acl}

\usepackage{times}
\usepackage{latexsym}
 \usepackage{graphicx}
\usepackage{booktabs}
\usepackage{enumitem}
 \usepackage[ruled,vlined]{algorithm2e}
\usepackage{multirow}
\usepackage{amsmath}
\usepackage{listings}
\usepackage{amsmath} % for mathematical features
\usepackage{amsfonts} % for mathematical fonts
\usepackage{colortbl} % для изменения цвета фона
\usepackage{hhline}   % для создания двойных линий
\usepackage{array}    % для настройки размера текста
\usepackage{xcolor}
\usepackage{colortbl}
\usepackage[utf8]{inputenc}
\usepackage{booktabs}
\usepackage{array}
\usepackage{tcolorbox}
\usepackage{subcaption}

\usepackage{mdframed}
\usepackage{subcaption} % для подфигур

\usepackage{xcolor}
\usepackage{subcaption}   % для подфигур
\usepackage{changepage}   % для изменения отступов с помощью adjustwidth
\usepackage{lipsum} 
\usepackage{caption}

\usepackage{graphicx}
\usepackage{float}

\usepackage[T1]{fontenc}
\usepackage[utf8]{inputenc}

\usepackage{pifont}
\usepackage{array}
\usepackage{caption}
\usepackage{subcaption}

\usepackage[colorinlistoftodos]{todonotes}

\definecolor{custombg}{HTML}{dff9fb} 
\definecolor{green_str}{HTML}{6ab04c}
\definecolor{code}{HTML}{130f40}

\usepackage{colortbl} 
\definecolor{soft_teal}{HTML}{2FBEAD} 
\definecolor{soft_tan}{HTML}{E8D5C4}

\newcommand{\bc}[1]{\cellcolor{soft_tan}{#1}}
\newcommand{\peachbox}[1]{\setlength{\fboxsep}{1pt}\colorbox{soft_tan!100}{#1}}

\definecolor{codegreen}{rgb}{0,0.6,0}
\definecolor{codegray}{rgb}{0.5,0.5,0.5}
\definecolor{codepurple}{rgb}{0.58,0,0.82}
\definecolor{backcolour}{rgb}{0.95,0.95,0.92}

\usepackage{microtype}
\usepackage{makecell}
\usepackage{xcolor}
\definecolor{lightblue}{RGB}{135,206,250}
\definecolor{lightyellow}{RGB}{255, 255, 224}

\definecolor{lightpink}{RGB}{255, 182, 193}

\definecolor{lightgray}{RGB}{211, 211, 211}

\definecolor{lightpurple}{RGB}{230, 230, 250}

\usepackage{arydshln}
\newboolean{showcomments}
\setboolean{showcomments}{true}

\ifthenelse{\boolean{showcomments}}{
  \newcommand{\nb}[3]{
    {\color{#2}\small\fbox{\bfseries\sffamily\scriptsize#1}}
    {\color{#2}\sffamily\small$\triangleright~$\textit{\small #3}$~\triangleleft$}
  }
}{
  \newcommand{\nb}[3]{}
}

\newcommand{\cm}{\ding{51}} % Create a shortcut for check mark
\newcommand{\xm}{\ding{55}} % Create a shortcut for cross mark

\title{CrafText Benchmark: Advancing Instruction Following in Complex Multimodal Open-Ended World}

\author{
Zoya Volovikova$^{1,2}$ \quad
Gregory Gorbov$^{2}$ \quad
Petr Kuderov$^{1,2}$ \\
\bf Aleksandr I. Panov$^{1,2,3}$ \quad
Alexey Skrynnik$^{1,2}$ \\
\\
$^1$AIRI, Moscow, Russia \\
$^2$MIPT, Moscow, Russia \\
$^3$FRC CSC RAS, Moscow, Russia \\
\texttt{volovikova@airi.net}
}
\begin{document}
\maketitle
\begin{abstract}

% Отметить про английские инструкции
Following instructions in real-world conditions requires a capability to adapt to the world's volatility and entanglement: the environment is dynamic and unpredictable, instructions can be linguistically complex with diverse vocabulary, and the number of possible goals an agent may encounter is vast. Despite extensive research in this area, most studies are conducted in static environments with simple instructions and a limited vocabulary, making it difficult to assess agent performance in more diverse and challenging settings. To address this gap, we introduce CrafText, a benchmark for evaluating instruction following in a multimodal environment with diverse instructions and dynamic interactions. CrafText includes 3,924 instructions with 3,423 unique words, covering Localization, Conditional, Building, and Achievement tasks. Additionally, we propose an evaluation protocol that measures an agent’s ability to generalize to novel instruction formulations and dynamically evolving task configurations, providing a rigorous test of both linguistic understanding and adaptive decision-making.

\end{abstract}

\section{Introduction}

Instruction following is a research area dedicated to developing methods that enable an agent to act within an environment based on natural language instructions and sensory input, such as visual observations.

%In real-world scenarios, the search space for finding an optimal decision-making strategy in instruction-following tasks expands due to the world's volatility and complex interdependencies. / making it more difficult to determine optimal strategies

\begin{figure}[!ht]
\centering
\includegraphics[width=\linewidth]{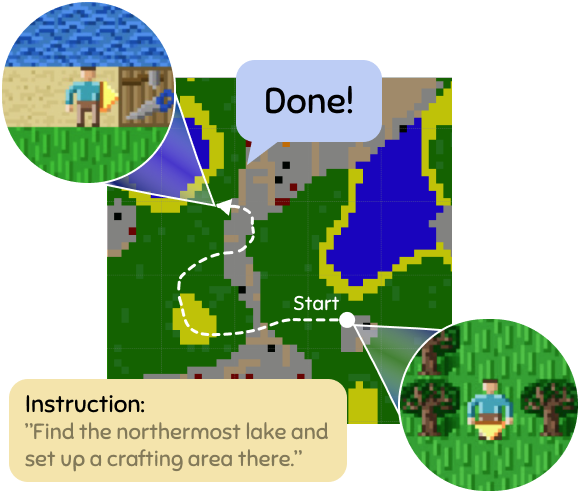}
\caption{An illustration depicting an agent navigating the CrafText environment to solve a task. The agent progresses from the starting point towards the northernmost lake, collecting the necessary resources along the way to set up a crafting table. A \textit{Done} marker indicates the location where the task is completed.}
\label{fig:visual_intro}
\end{figure}

In real-world scenarios, decision-making in instruction-following tasks becomes increasingly complex due to the world's volatility and intricate interdependencies. The environment is dynamic and unpredictable, requiring the agent to continuously adjust its decisions. Additionally, many instructions involve interactions with objects, and since the world contains a vast number of objects—each with multiple possible interactions—the space of potential goals further increases. Furthermore, instructions themselves can be phrased in multiple ways, making proper grounding essential: the agent must correctly interpret the instruction in terms of its intended goal and determine the specific objects and interactions required to achieve it.

Thus, an AI agent faces two primary challenges: (1) making decisions in a dynamically changing environment, and (2) generalizing its ability to follow instructions across diverse tasks and formulations by effectively linking natural language to observations. 

Individually, both challenges are well studied. Decision-making under dynamic and uncertain conditions in Reinforcement Learning (RL) has been explored in various studies~\cite{padakandla2021survey, attar2019reinforcement}. RL-based methods excel in open-world scenarios, where agents adapt to procedurally generated environments, evolving objects, and unpredictable interactions~\cite{hafner2021benchmarking, matthews2024craftax, guss2019minerl}. These approaches enable generalization, allowing agents to transfer learned policies to new settings and interact with unseen objects~\cite{stanic2023learning}.

Ways to connect different modalities, such as text and vision, can be found in studies on language grounding. Architectures like CLIP~\cite{yao2022detclip} and FILM~\cite{perez2018film} facilitate this connection by mapping textual instructions to visual inputs. These models have shown strong performance in tasks such as visual question answering (VQA)~\citep{de2023visual, ishmam2024image}, open-vocabulary segmentation, and image description~\citep{zhang2022dino, wang2305one, yao2022detclip}.

There are also studies that bridge language grounding and RL, addressing the problem of instruction following. These approaches leverage FiLM~\cite{zhong2019rtfm} and CLIP~\cite{paischer2023semantic, lynch2023interactive} to encode task-relevant features from both textual and visual modalities. However, these studies are constrained by the environments they rely on, which are often static or use procedurally generated instructions, limiting linguistic diversity. While environments like~\cite{Dynalang},~\cite{hanjie2021grounding}, and~\cite{zhong2019rtfm} introduce dynamics, their vocabulary remains restricted due to template-based instruction generation. Conversely,~\cite{shridhar2020alfred},~\cite{chen2020touchdownnaturallanguagenavigation} and~\cite{chen2020touchdownnaturallanguagenavigation} offer richer vocabularies but lack diverse interactions and environmental dynamics. Such settings narrow the search space for optimal policies, making it difficult to assess how well trained algorithms generalize to more dynamic and linguistically varied real-world conditions.  

%Moreover, as noted in~\citep{zhong2021silg}, the goals of instructions in these environments are often limited to a narrow set of tasks, such as object manipulation (lifting, lowering, moving), construction, or even simple navigation.

%Another key limitation of existing work is the lack of a robust evaluation framework for measuring how well agents generalize to new conditions. This includes assessing their ability to follow instructions with varied phrasing and complete tasks involving new goal configurations. It is also essential to test whether agents adapt when familiar objects appear in new contexts or when object-action associations change. 

In this paper, we introduce CrafText, a benchmark designed to evaluate an agent’s ability to follow complex natural language instructions in an interactive, multimodal environment. Unlike existing benchmarks, CrafText features an open-ended world dynamic with amount objects and way to interact with it, where objects change properties over time, requiring agents to continuously adapt and generalize beyond fixed action sequences.

We developed a linguistically diverse dataset with 3,924 instructions and 3,423 unique words to support instruction-following in this challenging setting. Tasks fall into four categories—Localization, Conditional, Building, and Achievement—evaluating an agent’s ability to interpret directions, follow conditions, construct structures, and achieve complex goals. Each instruction is paired with validation functions to systematically assess whether an RL agent successfully completes the specified goals.

% To support instruction-following in this challenging setting, we have developed a dataset that ensures linguistic diversity, consisting of 3,924 instructions and a rich vocabulary of 3,423 unique words. The tasks in the dataset are categorized into four distinct types—Localization, Conditional, Building, and Achievement—testing the agent’s ability to interpret directional instructions, follow conditional statements, construct various structures, and accomplish complex objectives. Additionally, for each instruction, we provide a set of validation functions that systematically assess whether an RL agent successfully completes the specified goals.

Additionally, we propose a specialized evaluation protocol to test the agent’s ability to generalize to novel instruction formulations and unseen goal configurations, providing a rigorous measure of both linguistic flexibility and adaptive decision-making.

To summarize, we make the following contributions:

% Shorter version
% \begin{itemize}
%     \item We introduce CrafText, a benchmark for training agents on goal-driven tasks with natural language instructions in a dynamic environment. It includes 3,924 instructions grouped into four categories—Localization, Conditional, Building, and Achievement—each paired with validation functions for precise task verification, and proposes an evaluation protocol that tests both linguistic adaptability and generalization to novel goal configurations.

%     \item We conduct a thorough evaluation, analyzing agent performance in CrafText using two well-known baselines: PPO with instruction embeddings as goal representations and Dynalang~\citep{Dynalang}, a specialized approach for multimodal tasks.

%     \item We implement CrafText as an open-source benchmark, providing its dataset and evaluation framework with support for XLA acceleration, enhancing computational efficiency and scalability. The dataset and code for CrafText are publicly available\footnote{\href{https://anonymous.4open.science/r/CrafText-D217/}{https://anonymous.4open.science/r/CrafText-D217/}}.
% \end{itemize}

\begin{itemize}
    \item We introduce CrafText, a benchmark for training agents on goal-driven tasks with natural language instructions in a dynamic environment.  

    \item We develop a dataset of 3,924 instructions, categorized into four distinct task types—Localization, Conditional, Building, and Achievement—along with validation functions for precise task verification in RL.  

    \item We propose an evaluation protocol that assesses both linguistic adaptability and generalization to novel goal configurations.  

    \item We conduct a thorough evaluation, analyzing agent performance in CrafText using two well-known baselines: PPO with instruction embeddings as goal representations and Dynalang~\citep{Dynalang}, a specialized approach for multimodal tasks.  
    % Not sura about open-sorce
    \item We implement CrafText as an open-source benchmark, providing its dataset and evaluation framework with support for XLA acceleration, enhancing computational efficiency and scalability. The dataset and code for CrafText are publicly available\footnote{\href{https://anonymous.4open.science/r/CrafText-D217/}{https://anonymous.4open.science/r/CrafText-D217/}}. 
    
\end{itemize}

\section{Related Work}

\begin{table*}[!ht]
\caption{This table provides a comparison of CrafText with several other multimodal environments across various characteristics, including vocabulary size, instruction length, benchmarking capabilities, stochastic transitions, world dynamics, the number of game objects, GPU acceleration, procedural world generation, and evaluation protocols. CrafText offers a well-balanced combination of features, supporting both stochasticity and dynamic world elements, with a significant number of game objects.}
\label{tab:envs-comparison}

\centering 
% \small
\rowcolors{2}{gray!15}{white}
\small
\begin{tabular}{ll >{\centering\arraybackslash}p{0.9cm} >{\centering\arraybackslash}p{0.9cm} >{\centering\arraybackslash}p{0.9cm} >{\centering\arraybackslash}p{0.9cm} >{\centering\arraybackslash}p{0.9cm} >{\centering\arraybackslash}p{0.9cm} >{\centering\arraybackslash}p{0.9cm} >{\centering\arraybackslash}p{0.9cm} >{\centering\arraybackslash}p{0.9cm}}
    \toprule
    \hiderowcolors % Disable alternating row colors
    Environment &
    \rotatebox{90}{\makecell{Repository}} &
    \rotatebox{90}{\makecell{Vocabulary \\ >3000 words}} & 
    \rotatebox{90}{\makecell{Maximum \\ Instruction \\ Length >50}} &
    \rotatebox{90}{\makecell{Benchmarking \\ Diverse \\ Abilities}} &
    \rotatebox{90}{\makecell{Stochastic \\ Transitions}} & 
    \rotatebox{90}{\makecell{Dynamic \\ World}} & 
    \rotatebox{90}{\makecell{Game \\ Objects >50}} & 
    \rotatebox{90}{\makecell{GPU \\ Accelerated}} & 
    \rotatebox{90}{\makecell{Procedural \\ Generated \\ World}} & 
    \rotatebox{90}{\makecell{Dual \\ Evaluation \\ Protocol}} \\
    \showrowcolors % Re-enable alternating row colors
    \midrule
    HomeGrid  & \href{https://github.com/jlin816/homegrid}{link}                                & \xm  & \xm & \xm & \xm & \cm & \xm & \xm & \cm & \xm \\
    BabyAI          & \href{https://github.com/mila-iqia/babyai}{link}                          & \xm  & \xm &  \xm & \xm & \cm & \xm & \xm & \cm & \xm \\
    RTFM            & \href{https://github.com/mlfoundations/rtfm/tree/main}{link}              & \xm  & \cm & \xm & \xm & \cm & \xm & \xm & \cm & \xm \\
    Messenger       & \href{https://github.com/ahjwang/messenger-emma}{link}     & \xm  & \cm & \xm & \xm & \cm & \xm & \xm & \cm & \xm \\
    Touchdown       & \href{https://github.com/lil-lab/touchdown}{link}                         & \cm  & \cm &  \xm & \xm & \xm & \xm & \xm & \cm & \xm \\
    Alfred          & \href{https://github.com/askforalfred/alfred}{link}                       & \cm  & \xm & \xm & \xm & \xm & \cm & \xm & \cm & \xm \\
    Cereal Bar       & \href{https://github.com/lil-lab/suhr2019executing}{link}                & \cm  & \xm &  \xm & \cm & \xm & \xm & \xm & \cm & \xm \\
    IGLU            & \href{https://github.com/iglu-contest/gridworld}{link} & \xm  & \cm & \xm & \cm & \xm & \xm & \xm & \xm & \xm \\
    MineDojo        & \href{https://github.com/MineDojo/MineDojo}{link}                         & \cm  & \cm & \cm & \cm & \cm & \cm & \xm & \cm & \xm \\
    CraftAssist        & \href{https://github.com/facebookresearch/craftassist}{link}           & \cm  & \cm & \xm & \cm & \cm & \cm & \xm & \cm & \xm \\
    CrafText (ours) & \href{https://anonymous.4open.science/r/CrafText-D217/}{link}             & \cm  & \cm & \cm & \cm & \cm & \cm & \cm & \cm & \cm \\
    \bottomrule
\end{tabular}

 \end{table*}

In this section, we provide an overview of existing multimodal environments and approaches for training agents in language-grounded decision-making. The focus is on highlighting the limitations of current environments in terms of versatility, dynamism, and language grounding, contrasted with the capabilities of CrafText, designed to address these gaps.

\textbf {Environments For Instruction Following.}

Our research focuses on developing benchmarks for training and evaluating agents in instruction-following tasks within dynamically changing environments, featuring diverse goals, interaction possibilities, and complex linguistic structures.

However, most existing environments are not well-suited for this research direction, as they are often designed for a specific task. For example, CraftAssist~\citep{gray2019craftassistframeworkdialogueenabledinteractive} and IGLU~\cite{kiseleva2023interactive} center around 3D construction but remain deterministic, preventing environmental changes from influencing decision-making. Similarly, Touchdown~\cite{chen2020touchdownnaturallanguagenavigation}, Alfred\cite{shridhar2020alfred} and CerealBar~\cite{suhr2019executing}, despite offering a rich instruction vocabulary, lack environmental dynamics and provide only a limited range of interactions with objects. 

%Alfred\cite{shridhar2020alfred}, designed for following instructions in a household environment, features numerous objects but restricts interactions to a small set of actions (Pickup, Put, Open, Close, ToggleOn, ToggleOff, Slice), reducing decision-making complexity.\bibliography{bib}

Dynamic environments such as HomeGrid~\cite{Dynalang}, BabyAI\cite{chevalier2018babyai}, RTFM\cite{zhong2019rtfm}, and Messenger~\cite{hanjie2021grounding} allow state changes, but their object interactions are limited, and their instructions are often procedurally generated, reducing linguistic diversity. More detailed information about each environment can be found in the Appendix \ref{app:env}

Moreover, none of the environments, including MineDojo~\cite{fan2022minedojo}, which allows configuring dynamic conditions, provide a dual evaluation protocol for analyzing both the ability to generalize to new linguistic constructions and the ability to generalize to new goals.

% Q:MOVE TO APPENDIX?
% In contrast to existing environments, CrafText presents an instruction-following challenge in a complex open-ended world: it features a dynamic environment, a large number of interactive objects with diverse interaction possibilities, and a rich set of linguistically varied instructions. This creates a vast decision space where the agent must learn to find an optimal policy. Additionally, the benchmark is fully implemented in JAX, utilizing XLA acceleration for efficient large-scale experimentation.

% Multimodal Decision Making
\textbf{Instruction Following Task} Language grounding in intelligent agents focuses on enabling agents to understand and execute textual instructions within virtual environments. Instruction following requires models to map natural language commands to appropriate actions, often integrating visual and textual information. Approaches to policy learning in multimodal environments can be categorized into three main groups. The first group utilizes CLIP~\cite{radford2021learning} to establish a shared representation space for textual and visual inputs, aiding in instruction grounding~\cite{paischer2023semantic, lynch2023interactive}. The second group employs projection layers~\cite{perez2018film, zhong2019rtfm, hanjie2021grounding} to align linguistic and perceptual features. The third category integrates transformer-based architectures, such as EmBERT~\cite{suglia2108embodied} and Vision-and-Language Navigation~\cite{savva2019habitat}, to process multimodal signals for task execution. Additionally, methods like those in~\cite{li2022grounded} and~\cite{brohan2023can} leverage transformers to improve instruction following. Finally, approaches such as Dynalang~\cite{Dynalang} employ cross-attention or hidden subspace compression to enhance the flexibility of text-conditioned policies, often within model-based reinforcement learning frameworks~\cite{hafner2023mastering}.

\section{Problem Statement}

The environment is formalized as a \textbf{goal-based Partially Observable Markov Decision Process (POMDP)}, represented by the tuple $(\mathcal{S}, \mathcal{A}, \mathcal{O}, \mathcal{T}, \mathcal{R}, \mathcal{G}, \gamma)$. This framework models the task of grounding natural language instructions to agent behavior in a partially observable and dynamic environment. The agent must interpret a provided textual instruction $I$, infer the underlying goal $g \in \mathcal{G}$, and execute actions to achieve it.

The state space $\mathcal{S}$ describes all possible configurations of the environment, while $\mathcal{A}$ defines the agent’s actions. Observations $o \in \mathcal{O}$ Lfprovide partial information about the current state and include both environmental data and the instruction $I$, which describes the desired outcome of the agent’s actions. The agent must use $I$ to recover a latent goal $g \in \mathcal{G}$ by approximating a grounding function $f_g(I)$ that predicts $g$ from the instruction. This grounding process forms the core of the task, requiring the agent to apply natural language processing techniques to interpret the instruction and link it to a meaningful goal.

The policy $\pi(a \mid o)$ maps the agent’s actions to observations $o$. The agent maintains a belief state $b(s)$, a probability distribution over possible states, to handle partial observability and guide decision-making under uncertainty. 

The objective is to find an optimal policy $\pi^*$ that maximizes the expected cumulative reward over time. Formally,

\[
\pi^* = \arg\max_{\pi} \mathbb{E}_\pi\left[\sum_{t=0}^{T} \gamma^t R(s_t, a_t, g) \mid o_0 \right].
\]

The environment introduces stochasticity through its transition dynamics $\mathcal{T}(s' \mid s, a)$, where the same action may lead to different outcomes. It is also dynamic, as states can evolve independently of the agent’s actions due to external factors like autonomous entity movement. These complexities demand a robust ability to interpret instructions, infer goals, and act effectively under uncertainty, forming the core challenges of this framework.

\section{CrafText}

\begin{figure}[!ht]
 \centering
 \includegraphics[width=0.65\linewidth]{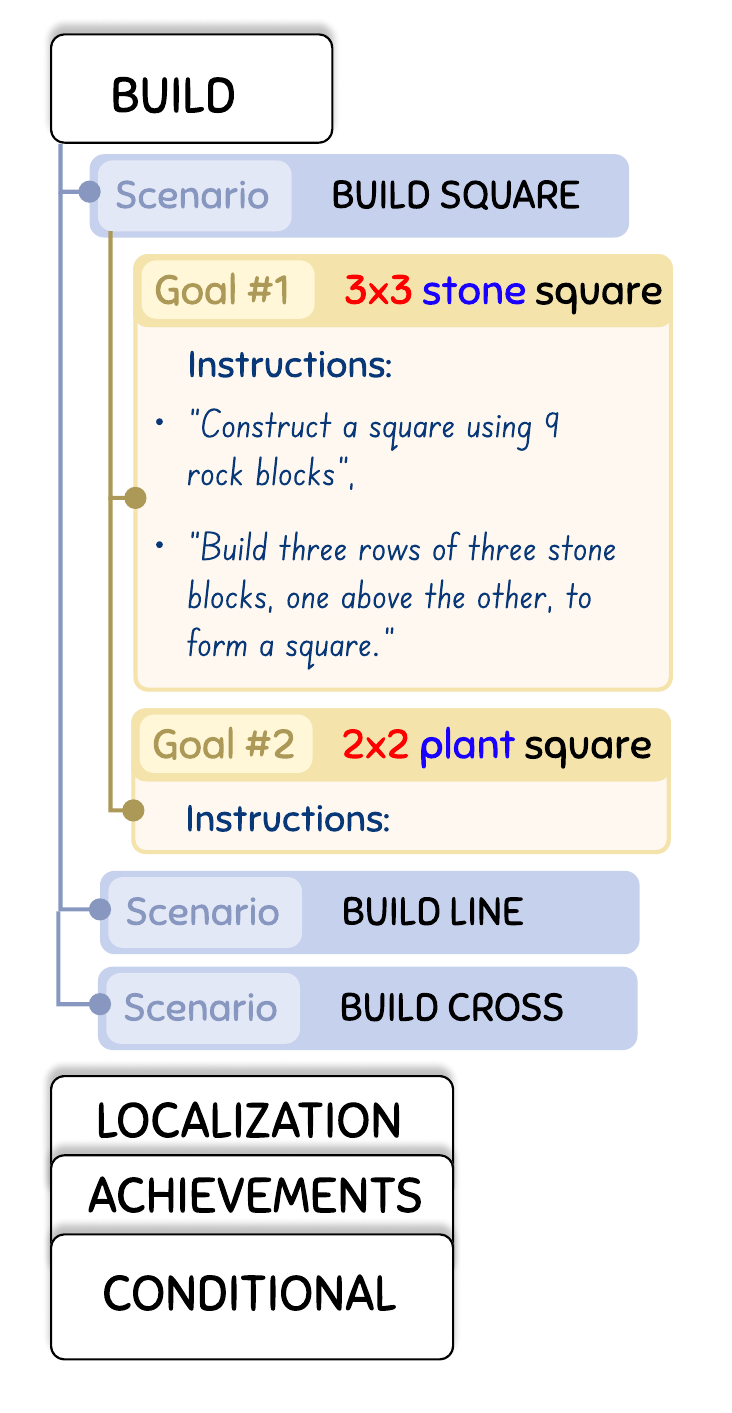}
 % \vspace*{-10px}
  \caption{The figure illustrates the hierarchical structure of the CrafText dataset. Each category contains multiple scenarios, each scenario includes different goals, and each goal is associated with multiple variations of instruction phrasing.}
 \label{fig:dataset_structure}
  %\vspace*{-18px}
\end{figure}

\begin{figure*}[!ht]
 \centering
 \includegraphics[width=1.0\linewidth]{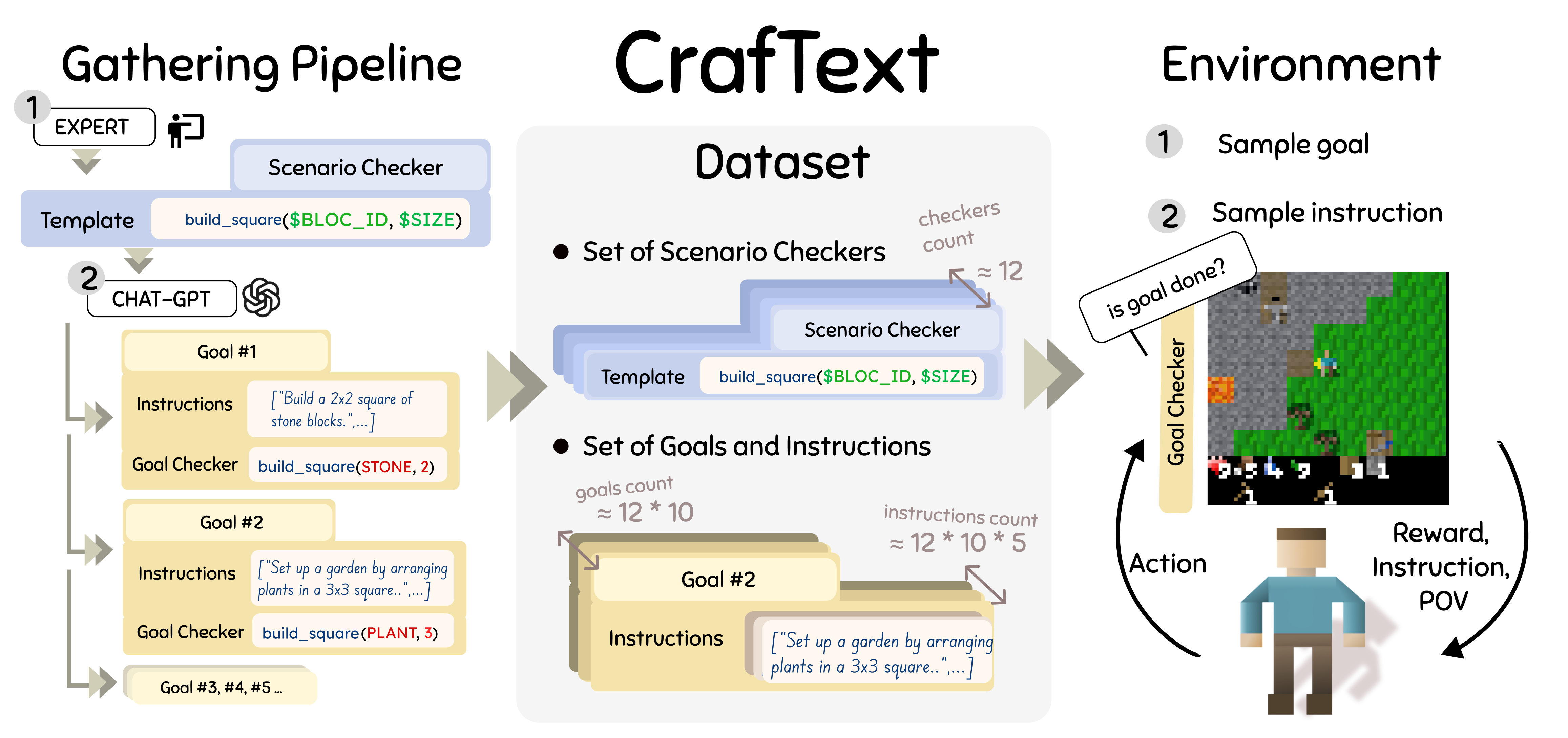}
 % \vspace*{-10px}
  \caption{{Left: \textit{Data Gathering Pipeline} -- experts define goal templates expanded with GPT to generate tasks, instructions, and goal-checking functions (e.g., build a square with stones or plants).
Middle: \textit{CrafText Dataset} -- features \textbf{162 goals} and \textbf{972 instructions} (162 $\times$ 6), combining scenario checkers, goals, and instructions with varied parameters like block type and size.
Right: \textit{Interactive Environment} -- the agent follows instructions and takes actions based on visual observations, while Goal Checkers verify progress by evaluating the state. The environment provides rewards and updates until the goal is achieved.}}
 \label{fig:data_overview}
  %\vspace*{-18px}
\end{figure*}

CrafText is a benchmark designed to evaluate an agent's ability to follow instructions in a complex, open-ended environment. The goal of the agent is to solve a diverse set of tasks, formulated in different ways in natural language, under dynamic conditions. The world is taken from Craftax~\cite{matthews2024craftax} (Appendix \ref{app:craftax}), providing a rich and interactive setting for agent evaluation.

As part of CrafText, we provide a \textbf{dataset} (see Section \ref{chapter:dataset}) that includes instructions from various task categories. The language used in these instructions is complex, featuring a large vocabulary. Each instruction is paired with a function that checks whether the agent has met the specified goal. Additionally, we introduce an \textbf{instruction generation framework} (Section \ref{chapter:gathering}), which was used to collect the data. This system allows for the expansion of the dataset by incorporating new tasks and scenarios while maintaining linguistic complexity.

We also present the \textbf{CrafText environment} (Section \ref{chapter:env}), which enables issuing new instructions to the agent within an episode and performing verification at each step. The implementation leverages JAX for high performance and computational speed, ensuring efficient training of agents in dynamic conditions (Appendix \ref{app:craftax}).

The data gathering pipeline, an overview of the dataset, and the agent-environment loop are illustrated in Figure~\ref{fig:data_overview}.

\subsection{Dataset: Overview and Structure} \label{chapter:dataset}

The CrafText dataset has a hierarchical structure (Figure \ref{fig:dataset_structure}), distinguishing three key concepts: \textit{Scenarios}, \textit{Goals}, and \textit{Instructions}. Scenarios represent abstract task classes that are not yet parameterized, such as building a square or placing an object at a certain distance from another. Goals are parameterized instances of these scenarios, specifying concrete details like building a 2x2 wooden square or placing a table to the left of a lake. Instructions are natural language variations of a goal’s formulation, providing multiple ways to express the same objective (see Appendix \ref{app:text} for examples).

To verify the completion of each instruction, we call a function that checks the execution of the scenario corresponding to the instruction with the given parameters. For example, for the instruction "Place a 2x2 wooden square," we call the Square Placement checker function with the parameters block = WOOD and side length = 2.

All scenario verification functions were implemented by a person familiar with the mechanics of the Craftax environment. In total, we have 12 scenarios across four task categories: \textit{Building}, \textit{Conditional}, \textit{Localization}, and \textit{Achievements}.

\textbf{Scenarios Categories}. \textit{Building} scenarios require the agent to construct a specified structure while remembering the starting point, since it may need to step away during construction to gather additional resources for placing blocks. In \textit{Conditional} scenarios, the agent’s understanding of instructions is tested with tasks such as "Craft a sword after gathering two stones," "Craft a sword before gathering two stones," or "Craft only a sword or only a pickaxe – whichever is needed for construction." \textit{Localization} scenarios evaluate the agent’s ability to correctly interpret spatial instructions, including compass directions (south, east, west, north) and relative directions (to the right, above, to the left, below) to accurately position objects relative to other map elements. Finally, \textit{Achievement} scenarios involve the agent performing standard in-game tasks—such as collecting wood, mining a diamond, or building a furnace—without additional complications in the instructions, sometimes requiring multiple achievements at once or the exclusion of certain actions. A more detailed breakdown of task categories can be found in the Appendix \ref{app:categories}, along with examples of instructions.

\textbf{Dataset Complexity Levels}.
In Craftax, which serves as the foundation for the benchmark, interacting with objects requires completing preliminary action sequences of varying lengths. For example, placing a stone involves the following steps: collecting wood → crafting a table → making a pickaxe → mining stone. Based on the length and complexity of these sequences, instructions are categorized into three difficulty levels: \textit{Easy}, \textit{Medium}, and \textit{Hard}. The Easy category includes tasks from the Achievement scenarios, assessing the agent’s ability to complete in-game achievements and their combinations. The Medium category covers all scenario types but is limited to tasks requiring relatively short action sequences (fewer than 10 steps). The Hard category includes tasks with either highly complex goals or long action sequences. A detailed list of task categories, their difficulty levels, and corresponding examples can be found in Appendix \ref{app:categories}.

\textbf{Test Dataset}. A key feature of our dataset is the inclusion of a dedicated test set designed to evaluate the agent's ability to generalize beyond the training distribution. The \textit{Paraphrased} subset assesses whether the agent can achieve the same goals encountered during training when instructions are reworded, allowing us to measure its ability to generalize across linguistic variations.

The \textit{New Objects  }subset evaluates the agent’s understanding of object properties by introducing new combinations of familiar objects. While all objects were present during training, they now appear in novel configurations. For example, if the training set included placing a stone block next to a lake and constructing a square from plantations, the test set might require constructing a square from stone blocks. If the agent succeeds, it demonstrates an understanding that stone blocks, like plantations, can be placed and that any placeable object can be used to construct geometric shapes such as squares. 

Examples of instructions in these test datasets can be found in Appendix \ref{app:text}.

%Together, these tests allow us to evaluate the agent’s ability to generalize both in terms of language comprehension and task execution in novel scenarios, ensuring a thorough assessment of its multimodal learning capabilities.

Dataset Size. The dataset consists of 12 scenarios with a total of 496 goals, 203 of which are reserved for testing. The goals are distributed based on complexity: 100 are classified as Easy (fully encompassing the Achievements category), 277 as Medium (which consists of multiple categories, including Achievements, leading to instruction overlap), and 219 as Hard. For each goal, we provide approximately 5–6 instructions, resulting in a total of around 3,924 instructions. The vocabulary size (unique word count) is 2,923. Detailed information is available in Appendix \ref{app:categories}.

\subsection{Instruction Generation Pipeline} 
\label{chapter:gathering}

In our benchmark, we aim to ensure both a substantial volume of textual instructions and a complex instructional language. Many existing approaches generate instructions by relying solely on procedural templates, resulting in limited linguistic diversity and a small vocabulary, despite a high instruction count. To address this, we introduce the \textit{Instruction Generation Pipeline}, which combines procedural goal generation for precise verification with large language models to achieve linguistic diversity.

The pipeline is centered around scenario checker functions implemented by our team, which encode the logic for goal verification. For each scenario checker, we define an acceptable range of parameters—such as block type, shape, and relative position—forming a reusable template for generating a large number of potential goals. For example, the template \texttt{BUILD\_\&SQUARE\&\_WITH\_\&STONE\&} includes the parameters \texttt{shape = square} and \texttt{material = stone}. By enumerating combinations of such parameters, we obtain a broad space of goals. A subset of these goals is selected for further refinement and instruction generation (see Appendix~\ref{app:gpt4prompr}).

At this stage, GPT-4 is used to generate natural language instructions and paraphrases corresponding to each selected goal, as well as the function call format needed to invoke the appropriate checker with the correct arguments. GPT-4 is prompted using a fixed, task-aware template (Appendix \ref{app:gpt4prompr}), and operates strictly as a language generator, without contributing to the underlying task logic or correctness checks.

\subsection{Environment}\label{chapter:env}

We developed the CrafText environment, an extension of Craftax that enables natural language instructions in the agent-environment loop. This extension is compatible with both \textit{Craftax-Classic} and the full version of \textit{Craftax}. The both possible observation types (visual and  vector-based) of \textit{Craftax} are augmented with instructions. 

\textbf{Episode Start}: At the start of each episode, the instruction and corresponding checker are randomly selected from the available options. It’s worth noting that a single checker can correspond to multiple instructions because it’s reusable through parameters. After that, the world is created, meaning that even for the same instruction, the environment can vary. 

\textbf{Reward System} The agent receives a reward of 1 for completing an instruction at the end of the episode. To verify completion, at each step, we run the corresponding Scenario Checker with goal-specific parameters. Additionally, we use the reward provided by the Craftax environment to incentivize discovering new achievements, scaling it down by a factor of 50.

\textbf{Episode Termination}  An episode concludes either when the step limit is reached (episode truncation), when the agent dies, or when the scenario checker function confirms that the instruction has been successfully completed. 

\textbf{JAX-acceleration} The entire code for the checker functions is implemented in \textit{JAX}, making the environment highly parallelizable (using JIT compilation) and allowing fast, large-scale training with GPU acceleration. % \todo[inline]{Add link to appedix Perfomance}

\section{Experiments}

% \begin{table*}[ht!]
% \footnotesize 
% \centering
% \caption{The success rates of PPO, the new baseline, and Dynalang were evaluated across 50 seeds, each representing a distinct world for every instruction within the complete set of \textbf{Medium Tasks}, where higher success rates indicate better performance. The best-performing approach is highlighted using \peachbox{tan boxes}.}
% \label{tab:ppo_dynalang_success}
% \setlength{\tabcolsep}{3.8pt} % Reduce column spacing
% \begin{tabular}{lccclccclcccl}
% \toprule
%  & \multicolumn{4}{c}{Train}& \multicolumn{4}{c}{Paraphrased}& \multicolumn{4}{c}{New objects}\\ 
% \cmidrule(lr){2-5} \cmidrule(lr){6-9} \cmidrule(lr){10-13}
% Algorithm       & PPO-T    & PPO-T+ & Dynalang   &FiLM& PPO-T        & PPO-T+    & Dynalang    &FiLM& PPO-T      & PPO-T+ & Dynalang  &FiLM\\ 
% \midrule
% Conditional    & 0.15    & \bc 0.17  & 0.0       &0.07& 0.12    & \bc  0.16& 0.0       &0.1& 0.12   & \bc 0.2  & 0.00      &0.17\\
% Build          &  0.25& 0.24  & 0.12      &\bc 0.38& 0.13    & 0.17& 0.09      &\bc 0.2& 0.13   &  0.17& 0.09      &\bc 0.2\\
% Localization   & \bc 0.33& 0.3      & 0.15      &0.29& \bc 0.35    & 0.3   & 0.13      &0.3& 0.17   & \bc 0.19  & 0.09      &0.19\\
% Achievements   & 0.55     &  0.7& 0.17      &\bc 0.76&  0.5&  0.48   & 0.1      &\bc 0.53& 0.34   & \bc 0.43  & 0.14      &0.38\\
% \midrule
% Total          &  0.4  & \bc 0.45  & 0.15      &0.43& \bc 0.36   &  0.35   & 0.05      &0.35& 0.22   & \bc 0.28  & 0.1      &0.26\\
% \bottomrule
% \end{tabular}
% \end{table*}
% \textbf{}

\begin{table*}[ht!]

\centering
\caption{The success rates of PPO, PPO-T+, FiLM, and Dynalang evaluated across 50 seeds, each representing a distinct world for every instruction within the complete set of \textbf{Medium Tasks}, where higher success rates indicate better performance. The best-performing approach is highlighted using \peachbox{tan boxes}. (Inverted)}
\label{tab:ppo_dynalang_success}
\begin{tabular}{ll>{\centering\arraybackslash}p{1.8cm}>{\centering\arraybackslash}p{1.8cm}>{\centering\arraybackslash}p{1.8cm}>{\centering\arraybackslash}p{1.8cm}>{\centering\arraybackslash}p{1.8cm}}
\toprule
 & Algorithm  
 & \begin{minipage}[c]{1.8cm}\centering\includegraphics[height=0.8cm]{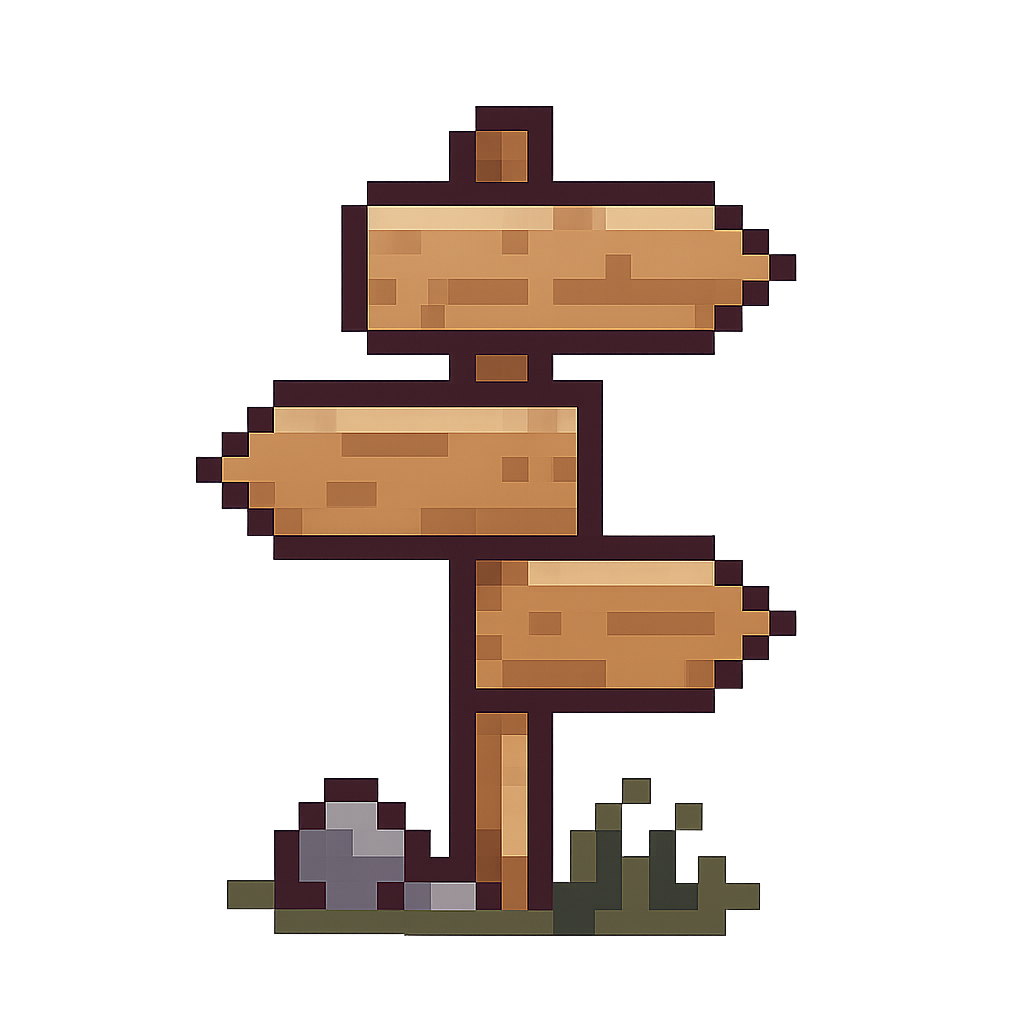}\newline Conditional \end{minipage}
 & \begin{minipage}[c]{1.8cm}\centering\includegraphics[height=0.8cm]{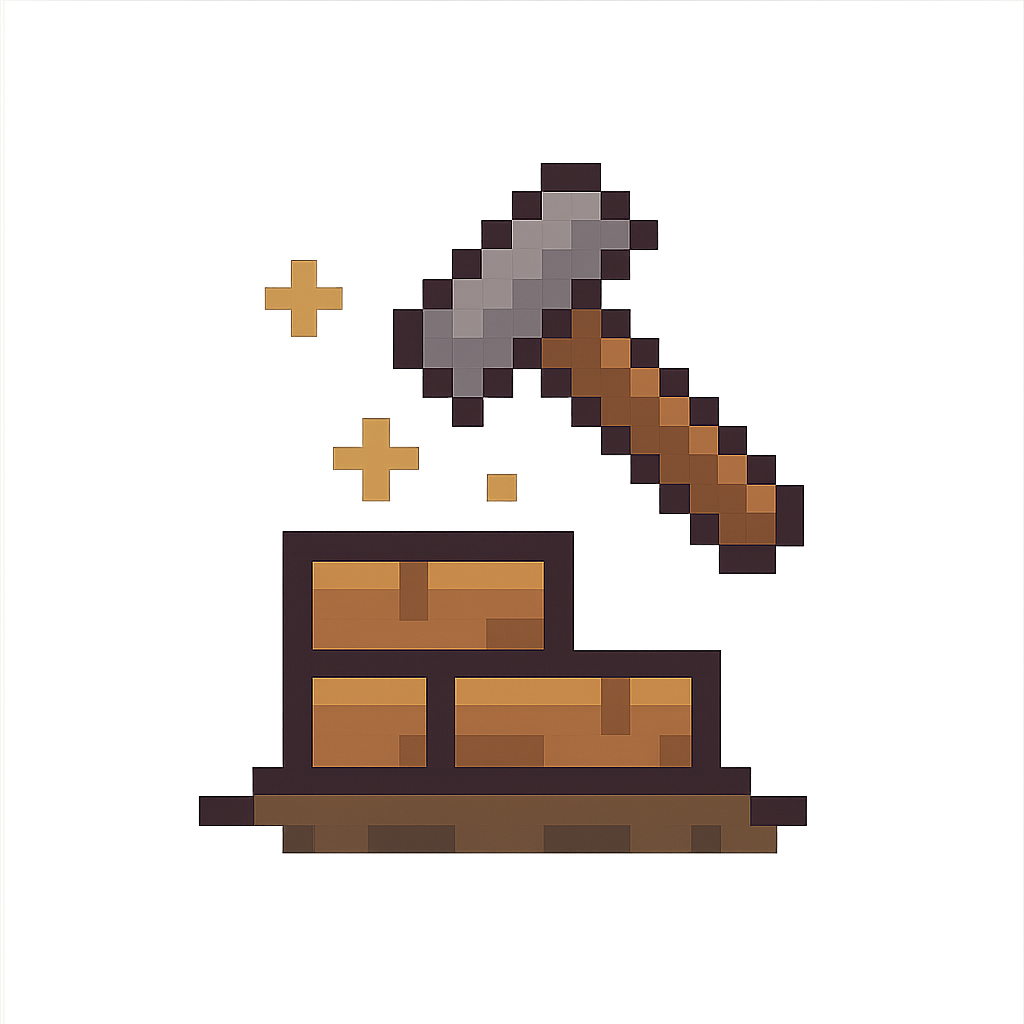}\newline Build \end{minipage}
 & \begin{minipage}[c]{1.8cm}\centering\includegraphics[height=0.8cm]{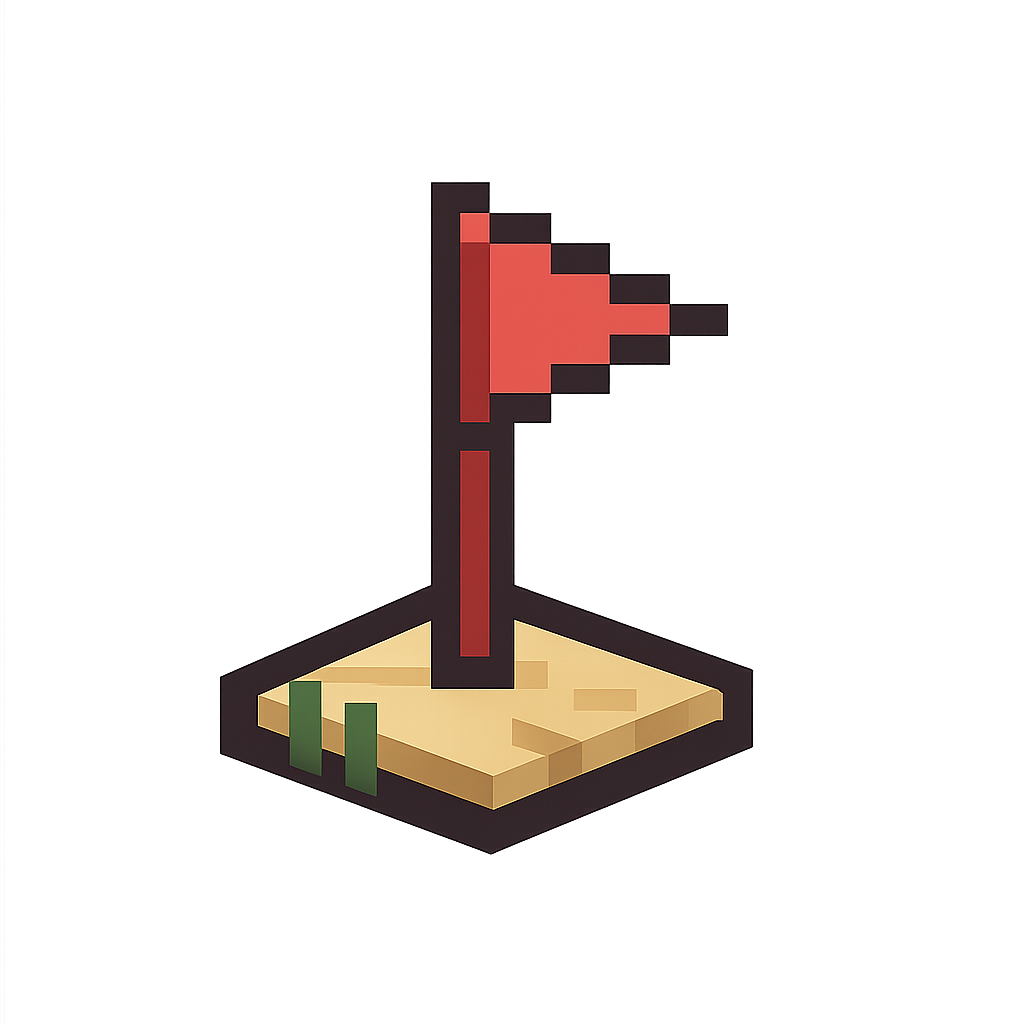}\newline Localization \end{minipage}
 & \begin{minipage}[c]{1.8cm}\centering\includegraphics[height=0.8cm]{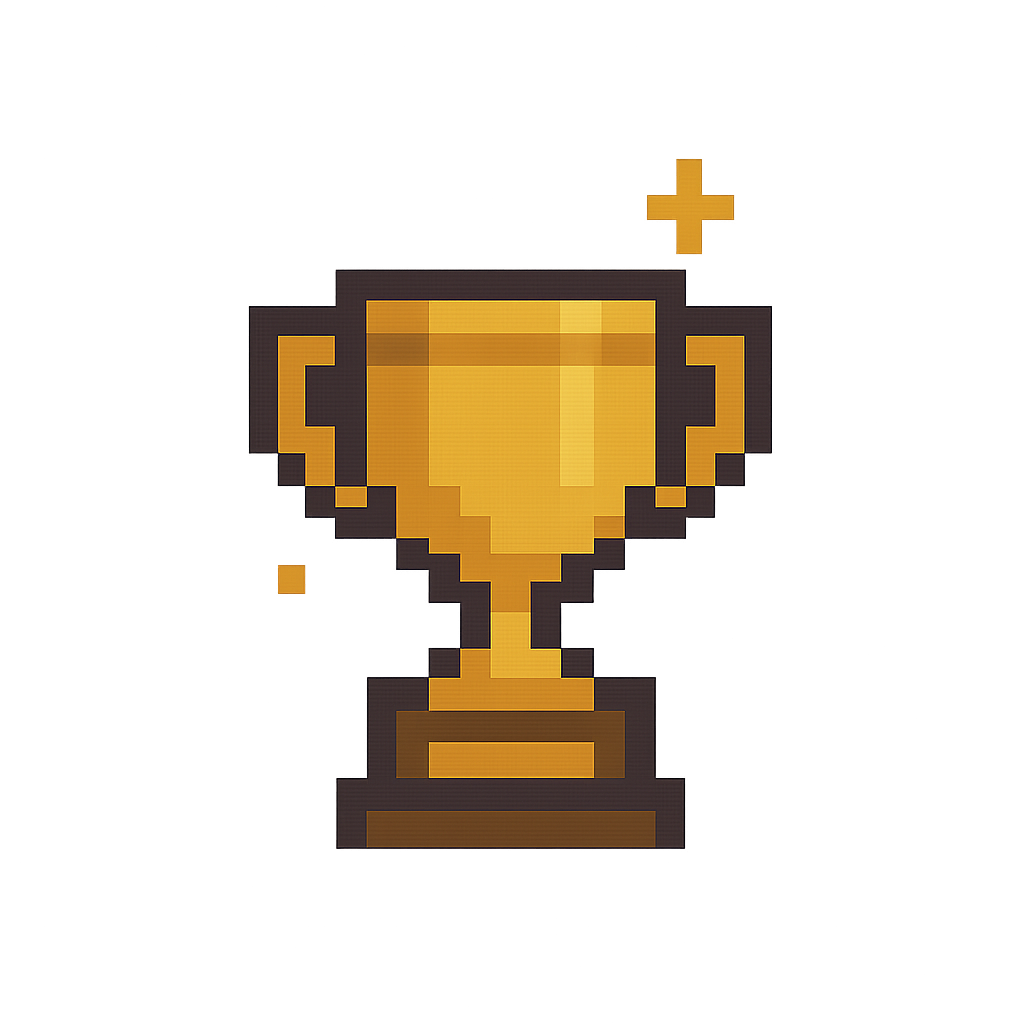}\newline Achievements \end{minipage}
 & Total \\
\midrule
\multirow{4}{*}{Train}        & PPO-T      & 0.15          & 0.25          & \bc 0.33     & 0.55          & 0.40         \\
                              & PPO-T+     & \bc 0.17     & 0.24          & 0.30         & 0.70          & \bc 0.45     \\
                              & Dynalang   & 0.00          & 0.12          & 0.15         & 0.17          & 0.15         \\
                              & FiLM       & 0.07          & \bc 0.38     & 0.29         & \bc 0.76     & 0.43         \\
\midrule
\multirow{4}{*}{Paraphrased}  & PPO-T      & 0.12          & 0.13          & \bc 0.35     & 0.50          & \bc 0.36     \\
                              & PPO-T+     & \bc 0.16     & 0.17          & 0.30         & 0.48          & 0.35         \\
                              & Dynalang   & 0.00          & 0.09          & 0.13         & 0.10          & 0.05         \\
                              & FiLM       & 0.10          & \bc 0.20     & 0.30         & \bc 0.53     & 0.35         \\
\midrule
\multirow{4}{*}{New objects}  & PPO-T      & 0.12          & 0.13          & 0.17         & 0.34          & 0.22         \\
                              & PPO-T+     & \bc 0.20     & 0.17          & \bc 0.19    & \bc 0.43     & \bc 0.28    \\
                              & Dynalang   & 0.00          & 0.09          & 0.09         & 0.14          & 0.10         \\
                              & FiLM       & 0.17          & \bc 0.20     & 0.19         & 0.38          & 0.26         \\
\bottomrule
\end{tabular}
\end{table*}

In this section, we address the following research questions (RQs) through comprehensive experiments:

\textbf{RQ1. (Task Handling in Dynamic Environments)}: How difficult is it for the agent to manage various categories of tasks in dynamic conditions?

\textbf{RQ2. (Generalization to New Instructions and Goals)}: Can the agent generalize its behavior:
   \begin{itemize}
      \item To paraphrased instructions that define the same goals as in the training set but are presented differently (Test Paraphrased)?
     \item To new goals that require solving familiar tasks but involve different combinations of objects and interactions not seen during training (Test New Objects)?
   \end{itemize}

To address these questions, we train several methods in the Medium part of our dataset and evaluate them using Success Rate (SR) as the primary metric. SR measures the proportion of episodes in which the agent successfully completes the given instruction. Each method was validated on two test datasets: \textit{Test Paraphrased} and \textit{Test New Objects}, as described in Section \ref{chapter:dataset}. To ensure a fair comparison, all models were trained for 48 hours on a Tesla V100 GPU. Below, we describe the baselines used in our evaluation.

% Short version
\textbf{PPO-T (Text-Augmented PPO).}
This method builds on the PPO~\cite{schulman2017proximal} implementation from the Craftax Baseline, using identical hyperparameters (Appendix \ref{app:ppo_params}). To enable instruction understanding, we integrate frozen DistilBERT embeddings by extracting the [CLS] token from the final hidden layer as a high-level representation. This embedding is concatenated with visual features to form a joint textual-visual representation. A GRU module is included to maintain contextual memory across time steps, enhancing the agent's ability to model sequential dependencies.

\textbf{PPO-T+ (Plan-Augmented PPO).}
To promote generalization, PPO-T+ extends PPO-T by introducing a planning step: each instruction is converted into a structured plan using GPT-4 (Appendix \ref{app:gpt4prompr}). The agent then follows the same training procedure as PPO-T, utilizing both visual observations and the GPT-generated plans to inform decision-making.

\textbf{FiLM.}
This baseline adopts PPO as the optimization method and integrates language through FiLM layers~\cite{perez2018film}, which apply feature-wise affine transformations to visual features conditioned on text. In our setup, transformation parameters are computed from the instruction and modulate each convolutional layer in the visual encoder, enabling instruction-aware visual processing.

\textbf{Dynalang.}
Dynalang~\cite{Dynalang} is a model-based method built on DreamerV3, designed for grounding language in multimodal environments. Rather than predicting actions directly, it forecasts future textual and visual states. Instructions are processed via the \textbf{T5} tokenizer, with optional embeddings. This approach achieves strong results in environments such as Messenger and BabyAI, demonstrating its effectiveness in learning predictive, grounded representations.

\subsection{Task Handling in Dynamic Environments (RQ1)}

Despite extensive training under optimal conditions, all tested algorithms exhibit suboptimal performance on the training set. Dynalang, achieves only a \(0.15\) success rate (SR), significantly underperforming across individual tasks. This result is particularly notable given DreamerV3's strong performance in environments like Crafter, which shares similarities with Craftax. One possible explanation is that the combination of complex textual instructions and the dynamic nature of the environment makes the learning process significantly more difficult. 

On the other hand, PPO-T, PPO-T+ and FiLM, which incorporate BERT-based instruction encoding without additional modifications, outperform Dynalang but still achieve only moderate success rates (\(\text{SR}\): PPO-T = \(0.40\), PPO-T+ = \(0.45\), FiLM = \(0.43 \)). This suggests that even with stronger language representations, existing methods struggle to generalize effectively in dynamic conditions, highlighting the need for further improvements in instruction-following capabilities.
%\todo[inline]{It is also possible to add an analysis in the Achievement Category because the results are lower than in Craftex PPO+RNN (with text instructions, max 0.7) and without it in ~Craftax (0.8+).}

\subsection{Generalization to New Instructions and Goals (RQ2)} 

We observe a general drop in performance on the \textit{Paraphrased} test set across all models. For PPO-T, PPO-T+, Dynalang, and FiLM, the Success Rate (SR) falls from $0.40$, $0.45$, $0.15$, and $0.43$ on the training set to $0.36$, $0.35$, $0.05$, and $0.35$, respectively.

Interestingly, PPO-T+ suffers the most from paraphrasing, with a drop of $0.10$, despite achieving the highest SR on the training set. This suggests that reformulated instructions may lead it to construct execution plans that diverge significantly from those encountered during training, thereby reducing its robustness to linguistic variation.

At the same time, PPO-T+ achieves the highest SR on the test set with entirely new goals \textit{New Objects} ($\text{SR} = 0.28$), outperforming FiLM ($0.26$), PPO-T ($0.22$), and Dynalang ($0.10$). This indicates that PPO-T+ is better at decomposing novel instructions into reusable subtasks, even when surface forms are unfamiliar — a sign of effective goal-level generalization.

FiLM comes closest to PPO-T+ in this setting, achieving competitive results without the explicit intermediate-step inference mechanism used by PPO-T+. The strong performance of FiLM may stem from its architectural design: FiLM layers in the encoder allow for more flexible integration of textual and visual information, which proves beneficial in handling novel object configurations.

\section{Conclusion}
We introduced CrafText, a benchmark for studying instruction following in dynamic environments with diverse objects, interactions, and high-vocabulary textual instructions. Agents must solve tasks across multiple categories, including Construction, Localization, Conditional tasks, and multi-step Achievement-based challenges.We show that agents that performed well in static environments with limited vocabulary, such as Dynalang, struggle with complex linguistic constructions and fail to generalize in dynamic environments. Our benchmark highlights the advantages of planning-based approaches, demonstrating that preprocessing instructions aids in solving novel tasks.

\section{Limitation}

The main limitation of this study is the absence of human-generated instructions in the dataset. While AI-generated instructions offer consistency and scalability, they may not capture the depth and nuance of human-crafted instructions, potentially limiting the model’s ability to generalize to complex, context-rich tasks common in real-world applications. Although ChatGPT-like models enhance accessibility, future work will focus on enriching the dataset with human input.

Another limitation is the lack of real-world interactive elements, where instructions are part of a dynamic conversation involving negotiation, clarification, collaboration, and adaptation. Expanding the benchmark to include such interactions would provide a more comprehensive evaluation of an AI’s ability to handle nuanced, real-life scenarios.

\bibliography{bib}

\begin{thebibliography}{32}
\expandafter\ifx\csname natexlab\endcsname\relax\def\natexlab#1{#1}\fi

\bibitem[{Attar and Dabirian(2019)}]{attar2019reinforcement}
Mehran Attar and Mohammadreza Dabirian. 2019.
\newblock Reinforcement learning for learning of dynamical systems in uncertain environment: A tutorial.
\newblock \emph{arXiv preprint arXiv:1905.07727}.

\bibitem[{Brohan et~al.(2023)Brohan, Chebotar, Finn, Hausman, Herzog, Ho, Ibarz, Irpan, Jang, Julian et~al.}]{brohan2023can}
Anthony Brohan, Yevgen Chebotar, Chelsea Finn, Karol Hausman, Alexander Herzog, Daniel Ho, Julian Ibarz, Alex Irpan, Eric Jang, Ryan Julian, et~al. 2023.
\newblock Do as i can, not as i say: Grounding language in robotic affordances.
\newblock In \emph{Conference on robot learning}, pages 287--318. PMLR.

\bibitem[{Chen et~al.(2020)Chen, Suhr, Misra, Snavely, and Artzi}]{chen2020touchdownnaturallanguagenavigation}
Howard Chen, Alane Suhr, Dipendra Misra, Noah Snavely, and Yoav Artzi. 2020.
\newblock \href {http://arxiv.org/abs/1811.12354} {Touchdown: Natural language navigation and spatial reasoning in visual street environments}.

\bibitem[{Chevalier-Boisvert et~al.(2018)Chevalier-Boisvert, Bahdanau, Lahlou, Willems, Saharia, Nguyen, and Bengio}]{chevalier2018babyai}
Maxime Chevalier-Boisvert, Dzmitry Bahdanau, Salem Lahlou, Lucas Willems, Chitwan Saharia, Thien~Huu Nguyen, and Yoshua Bengio. 2018.
\newblock Babyai: A platform to study the sample efficiency of grounded language learning.
\newblock \emph{arXiv preprint arXiv:1810.08272}.

\bibitem[{de~Faria et~al.(2023)de~Faria, Bastos, da~Silva, Fabris, Uchoa, Neto, and Santos}]{de2023visual}
Ana Cl{\'a}udia Akemi~Matsuki de~Faria, Felype de~Castro Bastos, Jos{\'e} Victor Nogueira~Alves da~Silva, Vitor~Lopes Fabris, Valeska de~Sousa Uchoa, D{\'e}cio Gon{\c{c}}alves de~Aguiar Neto, and Claudio Filipi Goncalves~dos Santos. 2023.
\newblock Visual question answering: A survey on techniques and common trends in recent literature.
\newblock \emph{arXiv preprint arXiv:2305.11033}.

\bibitem[{Fan et~al.(2022)Fan, Wang, Jiang, Mandlekar, Yang, Zhu, Tang, Huang, Zhu, and Anandkumar}]{fan2022minedojo}
Linxi Fan, Guanzhi Wang, Yunfan Jiang, Ajay Mandlekar, Yuncong Yang, Haoyi Zhu, Andrew Tang, De-An Huang, Yuke Zhu, and Anima Anandkumar. 2022.
\newblock \href {https://openreview.net/forum?id=rc8o_j8I8PX} {Minedojo: Building open-ended embodied agents with internet-scale knowledge}.
\newblock In \emph{Thirty-sixth Conference on Neural Information Processing Systems Datasets and Benchmarks Track}.

\bibitem[{Gray et~al.(2019)Gray, Srinet, Jernite, Yu, Chen, Guo, Goyal, Zitnick, and Szlam}]{gray2019craftassistframeworkdialogueenabledinteractive}
Jonathan Gray, Kavya Srinet, Yacine Jernite, Haonan Yu, Zhuoyuan Chen, Demi Guo, Siddharth Goyal, C.~Lawrence Zitnick, and Arthur Szlam. 2019.
\newblock \href {http://arxiv.org/abs/1907.08584} {Craftassist: A framework for dialogue-enabled interactive agents}.

\bibitem[{Guss et~al.(2019)Guss, Houghton, Topin, Wang, Codel, Veloso, and Salakhutdinov}]{guss2019minerl}
William~H Guss, Brandon Houghton, Nicholay Topin, Phillip Wang, Cayden Codel, Manuela Veloso, and Ruslan Salakhutdinov. 2019.
\newblock Minerl: A large-scale dataset of minecraft demonstrations.
\newblock \emph{arXiv preprint arXiv:1907.13440}.

\bibitem[{Hafner(2022)}]{hafner2021benchmarking}
Danijar Hafner. 2022.
\newblock \href {https://openreview.net/forum?id=1W0z96MFEoH} {Benchmarking the spectrum of agent capabilities}.
\newblock In \emph{The Tenth International Conference on Learning Representations, {ICLR} 2022, Virtual Event, April 25-29, 2022}. OpenReview.net.

\bibitem[{Hafner et~al.(2023)Hafner, Pasukonis, Ba, and Lillicrap}]{hafner2023mastering}
Danijar Hafner, Jurgis Pasukonis, Jimmy Ba, and Timothy Lillicrap. 2023.
\newblock Mastering diverse domains through world models.
\newblock \emph{arXiv preprint arXiv:2301.04104}.

\bibitem[{Ishmam et~al.(2024)Ishmam, Shovon, Mridha, and Dey}]{ishmam2024image}
Md~Farhan Ishmam, Md~Sakib~Hossain Shovon, Muhammad~Firoz Mridha, and Nilanjan Dey. 2024.
\newblock From image to language: A critical analysis of visual question answering (vqa) approaches, challenges, and opportunities.
\newblock \emph{Information Fusion}, page 102270.

\bibitem[{Kiseleva et~al.(2022)Kiseleva, Skrynnik, Zholus, Mohanty, Arabzadeh, C\^{o}t\'e, Aliannejadi, Teruel, Li, Burtsev, ter Hoeve, Volovikova, Panov, Sun, Srinet, Szlam, Awadallah, Rho, Kwon, Wontae~Nam, Bivort~Haiek, Zhang, Abdrazakov, Qingyam, Zhang, and Guo}]{kiseleva2023interactive}
Julia Kiseleva, Alexey Skrynnik, Artem Zholus, Shrestha Mohanty, Negar Arabzadeh, Marc-Alexandre C\^{o}t\'e, Mohammad Aliannejadi, Milagro Teruel, Ziming Li, Mikhail Burtsev, Maartje ter Hoeve, Zoya Volovikova, Aleksandr Panov, Yuxuan Sun, Kavya Srinet, Arthur Szlam, Ahmed Awadallah, Seungeun Rho, Taehwan Kwon, Daniel Wontae~Nam, Felipe Bivort~Haiek, Edwin Zhang, Linar Abdrazakov, Guo Qingyam, Jason Zhang, and Zhibin Guo. 2022.
\newblock \href {https://proceedings.mlr.press/v220/kiseleva23a.html} {Interactive grounded language understanding in a collaborative environment: Retrospective on iglu 2022 competition}.
\newblock In \emph{Proceedings of the NeurIPS 2022 Competitions Track}, volume 220 of \emph{Proceedings of Machine Learning Research}, pages 204--216. PMLR.

\bibitem[{Li et~al.(2022)Li, Zhang, Zhang, Yang, Li, Zhong, Wang, Yuan, Zhang, Hwang et~al.}]{li2022grounded}
Liunian~Harold Li, Pengchuan Zhang, Haotian Zhang, Jianwei Yang, Chunyuan Li, Yiwu Zhong, Lijuan Wang, Lu~Yuan, Lei Zhang, Jenq-Neng Hwang, et~al. 2022.
\newblock Grounded language-image pre-training.
\newblock In \emph{Proceedings of the IEEE/CVF Conference on Computer Vision and Pattern Recognition}, pages 10965--10975.

\bibitem[{Lin et~al.(2023)Lin, Du, Watkins, Hafner, Abbeel, Klein, and Dragan}]{Dynalang}
Jessy Lin, Yuqing Du, Olivia Watkins, Danijar Hafner, P.~Abbeel, Dan Klein, and Anca~D. Dragan. 2023.
\newblock \href {https://api.semanticscholar.org/CorpusID:260438420} {Learning to model the world with language}.
\newblock \emph{ArXiv}, abs/2308.01399.

\bibitem[{Lynch et~al.(2022)Lynch, Wahid, Tompson, Ding, Betker, Baruch, Armstrong, and Florence}]{lynch2023interactive}
Corey Lynch, Ayzaan Wahid, Jonathan Tompson, Tianli Ding, James Betker, Robert Baruch, Travis Armstrong, and Pete Florence. 2022.
\newblock \href {https://doi.org/10.48550/ARXIV.2210.06407} {Interactive language: Talking to robots in real time}.
\newblock \emph{CoRR}, abs/2210.06407.

\bibitem[{Matthews et~al.(2024)Matthews, Beukman, Ellis, Samvelyan, Jackson, Coward, and Foerster}]{matthews2024craftax}
Michael Matthews, Michael Beukman, Benjamin Ellis, Mikayel Samvelyan, Matthew Jackson, Samuel Coward, and Jakob Foerster. 2024.
\newblock Craftax: A lightning-fast benchmark for open-ended reinforcement learning.
\newblock In \emph{International Conference on Machine Learning ({ICML})}.

\bibitem[{Padakandla(2021)}]{padakandla2021survey}
Sindhu Padakandla. 2021.
\newblock A survey of reinforcement learning algorithms for dynamically varying environments.
\newblock \emph{ACM Computing Surveys (CSUR)}, 54(6):1--25.

\bibitem[{Paischer et~al.(2023)Paischer, Adler, Hofmarcher, and Hochreiter}]{paischer2023semantic}
Fabian Paischer, Thomas Adler, Markus Hofmarcher, and Sepp Hochreiter. 2023.
\newblock Semantic helm: A human-readable memory for reinforcement learning.
\newblock In \emph{Thirty-seventh Conference on Neural Information Processing Systems}.

\bibitem[{Perez et~al.(2018)Perez, Strub, De~Vries, Dumoulin, and Courville}]{perez2018film}
Ethan Perez, Florian Strub, Harm De~Vries, Vincent Dumoulin, and Aaron Courville. 2018.
\newblock Film: Visual reasoning with a general conditioning layer.
\newblock In \emph{Proceedings of the AAAI conference on artificial intelligence}, volume~32.

\bibitem[{Radford et~al.(2021)Radford, Kim, Hallacy, Ramesh, Goh, Agarwal, Sastry, Askell, Mishkin, Clark et~al.}]{radford2021learning}
Alec Radford, Jong~Wook Kim, Chris Hallacy, Aditya Ramesh, Gabriel Goh, Sandhini Agarwal, Girish Sastry, Amanda Askell, Pamela Mishkin, Jack Clark, et~al. 2021.
\newblock Learning transferable visual models from natural language supervision.
\newblock In \emph{International conference on machine learning}, pages 8748--8763. PMLR.

\bibitem[{Savva et~al.(2019)Savva, Malik, Parikh, Batra, Kadian, Maksymets, Zhao, Wijmans, Jain, Straub, Liu, and Koltun}]{savva2019habitat}
Manolis Savva, Jitendra Malik, Devi Parikh, Dhruv Batra, Abhishek Kadian, Oleksandr Maksymets, Yili Zhao, Erik Wijmans, Bhavana Jain, Julian Straub, Jia Liu, and Vladlen Koltun. 2019.
\newblock \href {https://doi.org/10.1109/ICCV.2019.00943} {Habitat: {A} platform for embodied {AI} research}.
\newblock In \emph{2019 {IEEE/CVF} International Conference on Computer Vision, {ICCV} 2019, Seoul, Korea (South), October 27 - November 2, 2019}, pages 9338--9346. {IEEE}.

\bibitem[{Schulman et~al.(2017)Schulman, Wolski, Dhariwal, Radford, and Klimov}]{schulman2017proximal}
John Schulman, Filip Wolski, Prafulla Dhariwal, Alec Radford, and Oleg Klimov. 2017.
\newblock Proximal policy optimization algorithms.
\newblock \emph{arXiv preprint arXiv:1707.06347}.

\bibitem[{Shridhar et~al.(2020)Shridhar, Thomason, Gordon, Bisk, Han, Mottaghi, Zettlemoyer, and Fox}]{shridhar2020alfred}
Mohit Shridhar, Jesse Thomason, Daniel Gordon, Yonatan Bisk, Winson Han, Roozbeh Mottaghi, Luke Zettlemoyer, and Dieter Fox. 2020.
\newblock \href {https://arxiv.org/abs/1912.01734} {{ALFRED: A Benchmark for Interpreting Grounded Instructions for Everyday Tasks}}.
\newblock In \emph{The IEEE Conference on Computer Vision and Pattern Recognition (CVPR)}.

\bibitem[{Stani{\'c} et~al.(2023)Stani{\'c}, Tang, Ha, and Schmidhuber}]{stanic2023learning}
Aleksandar Stani{\'c}, Yujin Tang, David Ha, and J{\"u}rgen Schmidhuber. 2023.
\newblock Learning to generalize with object-centric agents in the open world survival game crafter.
\newblock \emph{IEEE Transactions on Games}.

\bibitem[{Suglia et~al.(2021)Suglia, Gao, Thomason, Thattai, and Sukhatme}]{suglia2108embodied}
Alessandro Suglia, Qiaozi Gao, Jesse Thomason, Govind Thattai, and Gaurav~S. Sukhatme. 2021.
\newblock \href {https://api.semanticscholar.org/CorpusID:236975859} {Embodied bert: A transformer model for embodied, language-guided visual task completion}.
\newblock \emph{ArXiv}, abs/2108.04927.

\bibitem[{Suhr et~al.(2019)Suhr, Yan, Schluger, Yu, Khader, Mouallem, Zhang, and Artzi}]{suhr2019executing}
Alane Suhr, Claudia Yan, Charlotte Schluger, Stanley Yu, Hadi Khader, Marwa Mouallem, Iris Zhang, and Yoav Artzi. 2019.
\newblock Executing instructions in situated collaborative interactions.
\newblock \emph{arXiv preprint arXiv:1910.03655}.

\bibitem[{Wang et~al.(2023{\natexlab{a}})Wang, Xie, Jiang, Mandlekar, Xiao, Zhu, Fan, and Anandkumar}]{wang2023voyager}
Guanzhi Wang, Yuqi Xie, Yunfan Jiang, Ajay Mandlekar, Chaowei Xiao, Yuke Zhu, Linxi Fan, and Anima Anandkumar. 2023{\natexlab{a}}.
\newblock Voyager: An open-ended embodied agent with large language models.
\newblock \emph{arXiv preprint arXiv:2305.16291}.

\bibitem[{Wang and Narasimhan(2021)}]{hanjie2021grounding}
H.~J.~Austin Wang and Karthik Narasimhan. 2021.
\newblock \href {https://api.semanticscholar.org/CorpusID:231639188} {Grounding language to entities and dynamics for generalization in reinforcement learning}.
\newblock \emph{ArXiv}, abs/2101.07393.

\bibitem[{Wang et~al.(2023{\natexlab{b}})Wang, Wang, Lin, Bai, Zhou, Zhou, Wang, and Zhou}]{wang2305one}
Peng Wang, Shijie Wang, Junyang Lin, Shuai Bai, Xiaohuan Zhou, Jingren Zhou, Xinggang Wang, and Chang Zhou. 2023{\natexlab{b}}.
\newblock One-peace: Exploring one general representation model toward unlimited modalities.
\newblock \emph{arXiv preprint arXiv:2305.11172}.

\bibitem[{Yao et~al.(2022)Yao, Han, Wen, Liang, Xu, Zhang, Li, Xu, and Xu}]{yao2022detclip}
Lewei Yao, Jianhua Han, Youpeng Wen, Xiaodan Liang, Dan Xu, Wei Zhang, Zhenguo Li, Chunjing Xu, and Hang Xu. 2022.
\newblock Detclip: Dictionary-enriched visual-concept paralleled pre-training for open-world detection.
\newblock \emph{Advances in Neural Information Processing Systems}, 35:9125--9138.

\bibitem[{Zhang et~al.(2022)Zhang, Li, Liu, Zhang, Su, Zhu, Ni, and Shum}]{zhang2022dino}
Hao Zhang, Feng Li, Shilong Liu, Lei Zhang, Hang Su, Jun Zhu, Lionel~M Ni, and Heung-Yeung Shum. 2022.
\newblock Dino: Detr with improved denoising anchor boxes for end-to-end object detection.
\newblock \emph{arXiv preprint arXiv:2203.03605}.

\bibitem[{Zhong et~al.(2019)Zhong, Rockt{\"a}schel, and Grefenstette}]{zhong2019rtfm}
Victor Zhong, Tim Rockt{\"a}schel, and Edward Grefenstette. 2019.
\newblock Rtfm: Generalising to novel environment dynamics via reading.
\newblock \emph{arXiv preprint arXiv:1910.08210}.

\end{thebibliography}

\newpage

\appendix

%------------------------------------------------------------------------------------------------
\section{DATASET: Tasks Description} \label{app:categories}

\textbf{Conditional}. In the \textit{Conditional} category, the agent is required to understand the sequence of actions it needs to perform and the order in which these actions should be carried out. 

%%%%%%
\begin{figure}[H]
\footnotesize
\begin{mdframed}[backgroundcolor=gray!10, roundcorner=10pt, linewidth=0pt, frametitle={\textbf{Example}}, frametitlebackgroundcolor=gray!20, nobreak=true]
    \footnotesize
    \noindent
    \textit{Instruction:} "After collecting coal, the player should gather wood and then place a stone on the crafting table."
    \\

   % \textit{Expected Plan:} The agent needs to first identify the sequence of actions (collect coal, gather wood, place stone) and then execute them in the correct order.
\end{mdframed}
\vspace*{-12px}
\caption{Sequencing Instruction Example}
\label{fig:Sequencing}
\vspace*{-6px}
\end{figure}

%%%%%%
The focus here is on understanding temporal relationships between actions (see an example at Figure \ref{fig:Sequencing}). %The agent must recognize words like "after" to correctly sequence the actions. The common vocabulary includes terms like "player," "after," "coal," "place," and "stone," which are integral to these sequences.

\textbf{Build}. The \textit{Building} category involves tasks where the agent must construct specific shapes or structures based on verbal instructions.

%%%%%%%%
\begin{figure}[H]
\footnotesize
\begin{mdframed}[backgroundcolor=gray!10, roundcorner=10pt, linewidth=0pt, frametitle={\textbf{Example}}, frametitlebackgroundcolor=gray!20, nobreak=true]
    \footnotesize
    \noindent
    \textit{Instruction:} "Arrange the tables in a square pattern using 9 stone blocks."

\end{mdframed}
\vspace*{-12px}
\caption{Building Instruction Example}
\label{fig:building}

\end{figure}
%%%%%%%%%

This category tests the agent's spatial reasoning and ability to translate instructions into precise constructions (see an Example at Figure \ref{fig:building} ). 

\textbf{Localization}. In the Localization category, the agent must determine from which side it should position an object and in relation to which reference object.

%%%%%%%%%%%
\begin{figure}[H]
\footnotesize
\begin{mdframed}[backgroundcolor=gray!10, roundcorner=10pt, linewidth=0pt, frametitle={\textbf{Example}}, frametitlebackgroundcolor=gray!20, nobreak=true]
    \footnotesize
    \noindent
    \textit{Instruction:} "Place the table two steps to the left of the lake."
\end{mdframed}
\vspace*{-12px}
\caption{Localization Instruction Example}
\label{fig:localization}
\vspace*{-6px}
\end{figure}

%%%%%%%%%%%%%

In this example, the agent needs to: 1) Identify the lake 2) Determine the left side 3) Understand relative positioning Additionally, these tasks may involve directional terms such as left, right, above, below, north, south, west, and east.

This category is particularly challenging as it requires the agent to integrate directional instructions with map-based navigation (see an example at Figure \ref{fig:localization}.

\textbf{Achievements Tasks}. The \textit{Achievements} category contains tasks that require the agent to complete in-game achievements. It also includes tasks that instruct the agent to perform actions involving combinations of achievements (Figure \ref{fig:achievments}).
%%%%%%%%%%%
\begin{figure}[H]
\footnotesize
\begin{mdframed}[backgroundcolor=gray!10, roundcorner=10pt, linewidth=0pt, frametitle={\textbf{Examples}}, frametitlebackgroundcolor=gray!20, nobreak=true]
    \footnotesize
    \noindent
    \textit{Instruction:} "Forge a sturdy pickaxe from stone."

    \textit{Instruction:} "Craft a wooden pickaxe but avoid making a stone pickaxe."

     \textit{Instruction:} "Craft a stone sword and eliminate all undead."
\\

\end{mdframed}
\vspace*{-12px}
\caption{Achievements Tasks}
\label{fig:achievments}
\vspace*{-6px}
\end{figure}

This type of task evaluates the agent's ability to understand achievements by interpreting tasks and identifying the necessary actions. It also tests whether the agent can combine multiple achievements in sequence or simultaneously. Additionally, it assesses the agent's ability to follow constraints, ensuring specific conditions are met while completing a task.
%------------------------------------------------------------------------------------------------
\section{DATASET: Instructions Examples}
\label{app:text}

\begin{figure}[ht!]
    \centering
    \includegraphics[width=1.0\linewidth]{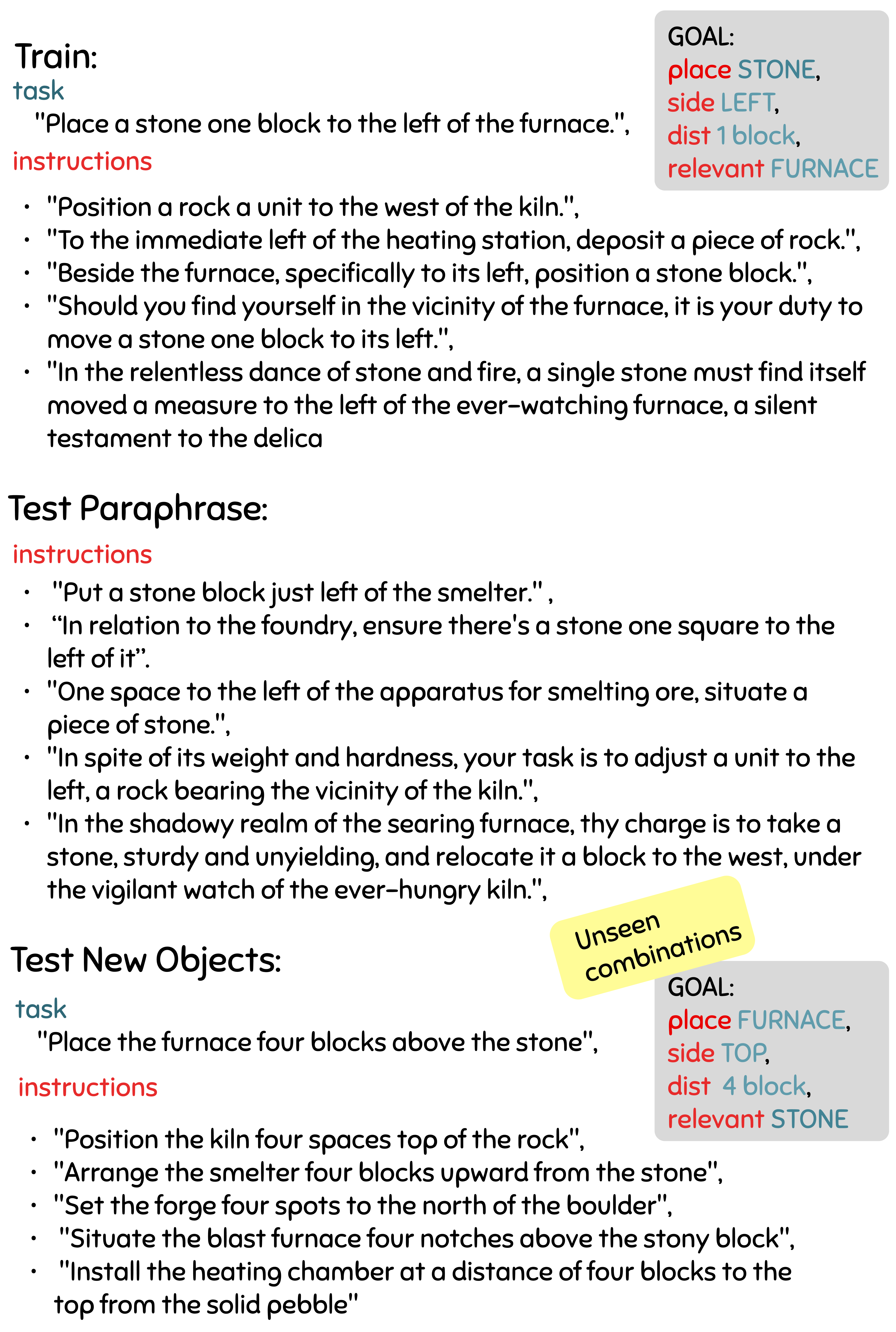}
    \caption{Examples of instructions in dataset.}
    \label{fig:localization_instr_example}
\end{figure}

\begin{figure*}[ht!]
    \centering
    \hspace*{-1.5cm}  
    \includegraphics[width=1.2\textwidth]{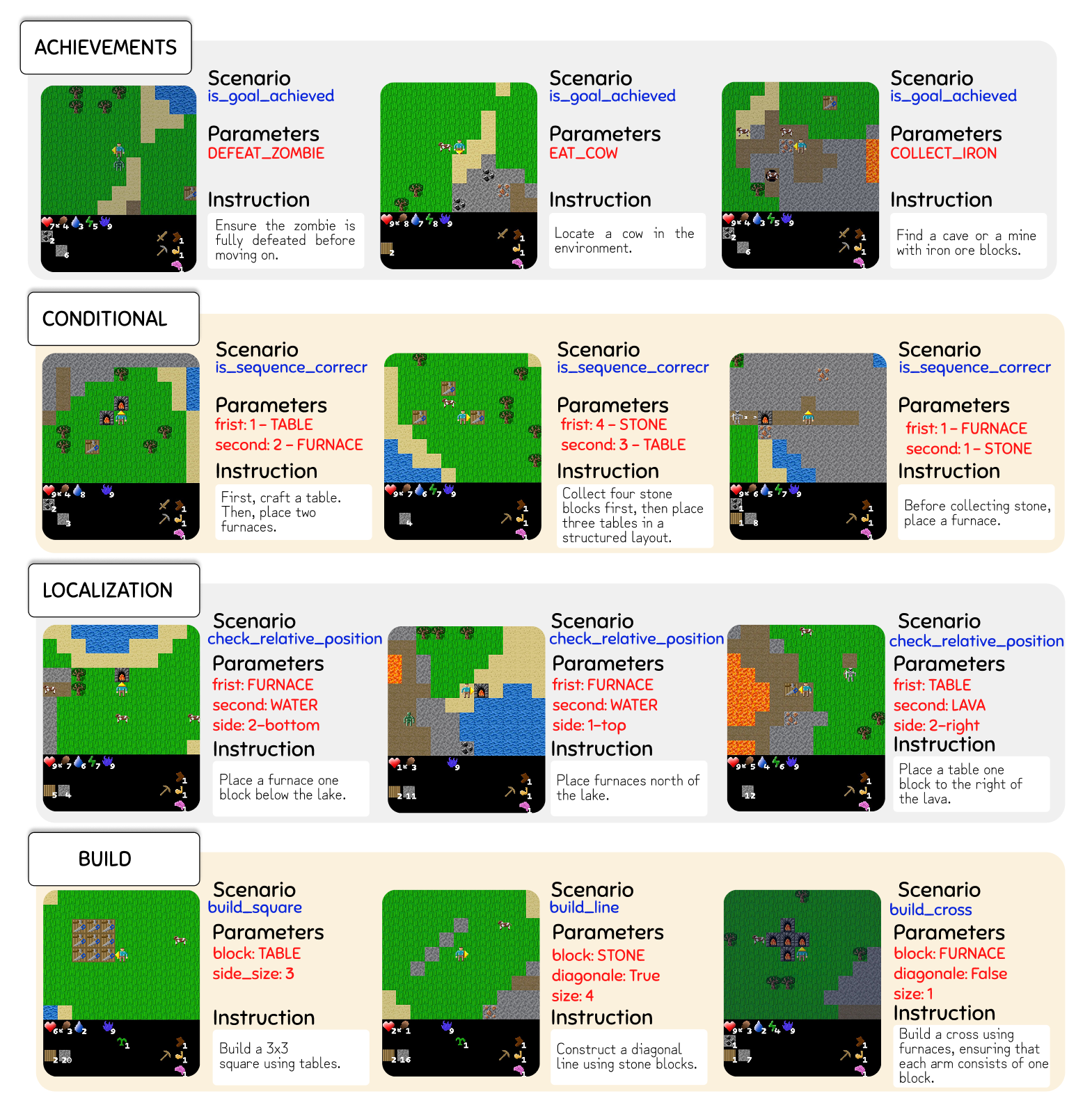}
    \caption{Example instruction set for different tasks.}
    \label{fig:instruction_examples}
\end{figure*}

\section{DATASET: Per Category Scenarios, Goal, Instructions}
\label{app:instructions_examples}

\begin{table*}[!ht]
\centering \footnotesize
\caption{Summary of Possible (Possible G) and Implemented Goals (Implemented G), with Complexity}
\label{tab:scalability}  % Correct placement of label
\begin{tabular}{l|l|l|l|l|l}

\textbf{Name}                                & \textbf{Class Name} & \textbf{Args} & \textbf{Possible G} & \textbf{Implemented G} & \textbf{Complexity} \\ \hline

achievements                                 & achievements       & list             & 500                      & 164                        & EASY \\ %\hline

build\_line                                  & build               & 3               & 75                      & 57                         & MEDIUM  \\ %\hline
square                                       & build               & 2               & 25                      & 28                         & MEDIUM \\ %\hline
cross                                        & build               & 3               & 75                      & 10                          & HARD \\ %\hline

localization\_place                          & localization        & 3                & 120                      & 91                         & MEDIUM \\ %\hline
water\_sources                               & localization        & 2                & 25                      & 15                         & HARD \\ %\hline

place\_item\_after\_collection               & conditional         & 3               & 80                      & 55                          & MEDIUM \\ %\hline
collect\_item\_after\_place                  & conditional         & 3               & 80                      & 56                          & MEDIUM \\ %\hline

build\_line\_and\_localization\_place            & combo               & 4            & 25 x 25                 & 5                          & HARD \\ %\hline
build\_line\_after\_collect                    & combo               & 4             & 25 x 25                 & 5                          & HARD \\ %\hline
collect\_item\_and\_build\_square               & combo               & 4            & 25 x 25                 & 5                          & HARD \\ %\hline
water\_sources\_and\_localization\_place        & combo               & 4            & 25 x 25                 & 5                          & HARD \\ \hline

\textbf{Total}                               &                    &                 & \textbf{3480}           & \textbf{496}                          & \\ \hline
\end{tabular}
\end{table*}

In our dataset, there is a division into the training set, the test set with rephrased instructions (\textit{Test Paraphrased}), and the test set with new tasks (\textit{Test New Object}). For example, in the localization scenario, where blocks need to be placed relative to other blocks at a certain distance, if the training set includes a task to place a stone block to the left of a furnace, then in \textit{Test Paraphrased}, it may be rephrased as "find the furnace and place a stone block one block north of it." In \textit{Test New Object}, a new combination of parameters appears, such as a task to place a furnace four blocks above a stone. The complexity of \textit{Test New Object} lies in the fact that the agent has multiple ways to approach the task. It can either find existing stone blocks and place the furnace relative to them or place both the stone and the furnace itself, as in the training task. This adds variability to the problem and requires the agent to be flexible in decision-making. See Figure \ref{fig:localization_instr_example} for more examples.

%------------------------------------------------------------------------------------------------

\section{DATASET: Generation Prompt}
\label{app:prompt}

The code for checking played is following:

\begin{lstlisting}[language=Python, basicstyle=\ttfamily\small, keywordstyle=\color{blue}]
 #Implementation of the scanerio checker
\end{lstlisting}
A scenario consists of instructions given by player 1 to player 2. Player 2 follows these instructions, which are then verified by a corresponding function. For the function scenario.py, please provide realistic examples of instructions that player 1 might give, along with 5 paraphrases for each.

Requirements:

1) When specifying target objects (objects with which the player will interact), use different synonyms in paraphrases to assess the vocabulary range of player 2.

2) Present the target objects in varying orders to evaluate how well player 2 understands different language structures.

3) For each set of paraphrases, sort them from the simplest language to the most complex.

4) Ensure the instructions are as varied as possible with a broad vocabulary.

Format your answer as a Python dictionary with the following structure:

\begin{lstlisting}[language=Python, basicstyle=\ttfamily\small, keywordstyle=\color{blue}]
instructions = {
    instruction_id: {
        'instruction': "Example instruction here",
        'instruction_paraphrases': [
            "Paraphrase 1 here",
            "Paraphrase 2 here",
            "Paraphrase 3 here",
            "Paraphrase 4 here",
            "Paraphrase 5 here"
        ],
        'check_lambda': lambda ...:scenario_function(...): ...  # Example usage of the function
    }
    }
\end{lstlisting}

%------------------------------------------------------------------------------------------------

In Figure~\ref{fig:instruction_examples}, we present examples of instructions along with their corresponding Goal Checker functions for each task category: BUILD, CONDITIONAL, and LOCALIZATION. The instructions guide the agent in performing specific actions within the environment, while the Goal Checker functions ensure that the goals of these tasks are met. We also provide visual illustrations from the termination states of each task, demonstrating the final result expected from the agent's actions.

%------------------------------------------------------------------------------------------------
\section{DATASET: Assessing the Scalability}
\label{app:scale}

Table \ref{tab:scalability} summarizes possible and implemented goals across different scenario classes and complexities. It lists the class, number of arguments (Args), possible goals, and implemented goals. Each instruction's complexity level (Easy, Medium, Hard) is also indicated. The dataset defines 3,480 possible goals, which expand to over 20,000 instructions when factoring in paraphrases.

Despite having 496 distinct goals, the dataset's scalability stems from a function-based task generation process. This approach systematically combines predefined goals, environmental parameters, and paraphrased instructions, creating diverse training interactions for instruction-following agents.

A key example is the \texttt{place} function, which checks whether an object is positioned relative to a target at a specified side (right, left, top, bottom) and distance. With four object types and four distances, this results in 512 unique goals. Each goal is paraphrased in six ways, producing 3,072 instructions from this function alone.

This scalability extends to other task-generating functions, enabling continuous dataset expansion. Appendix \ref{app:scale} details how this methodology scales to 3,480 unique goals and approximately 20,880 instructions with paraphrases.

%------------------------------------------------------------------------------------------------
\section{ENVIRONMENT: Perfomance} \label{app:sps}

Efficient simulation speed is a cornerstone of reinforcement learning (RL) environments, enabling faster training cycles and broader experimentation. CrafText, built on top of Craftax, leverages the computational power of JAX and advanced parallel execution to demonstrate remarkable scalability. As the number of parallel environments grows, CrafText's steps per second (SPS) increase proportionally, unlocking the potential for large-scale RL research and development.

To evaluate performance, we measured the environment's speed over \textbf{5000 steps} using an agent with \textbf{random behavior}. The experiments compared SPS across different base Craftax environments and hardware configurations, including devices such as \textbf{Tesla V100 GPUs}.  On Classic Craftax, training on 1024 environments achieved approximately 44,630 SPS, while on CrafText, the same configuration yielded around 9,100 SPS on a Tesla V100.

\begin{figure}[ht!]
    \centering
    \includegraphics[width=0.9\linewidth]{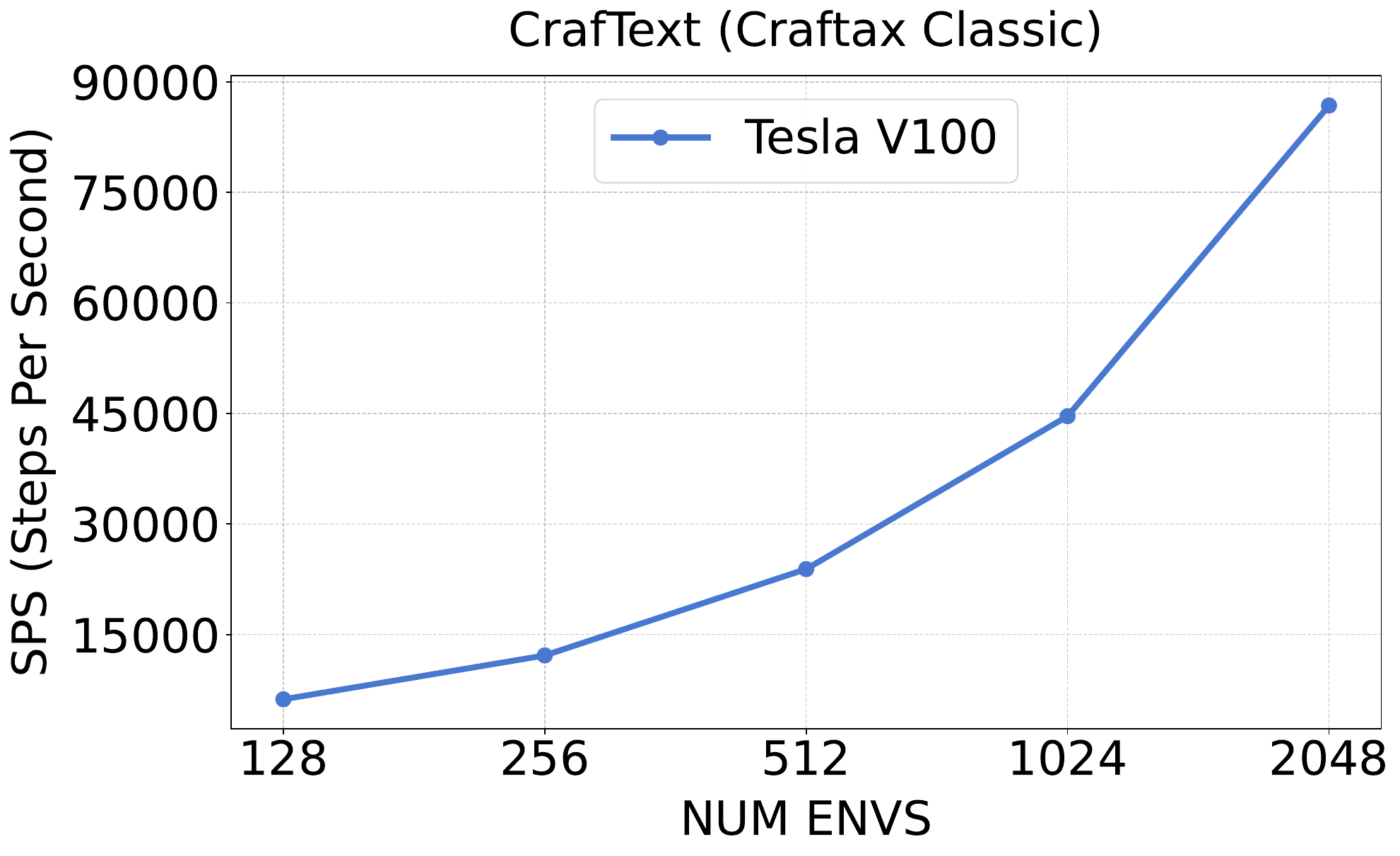}
    \caption{SPS of CrafText environment based on Craftax Classic environment}
    \label{fig:sps_classic}
\end{figure}

\begin{figure}[ht!]
    \centering
    \includegraphics[width=0.9\linewidth]{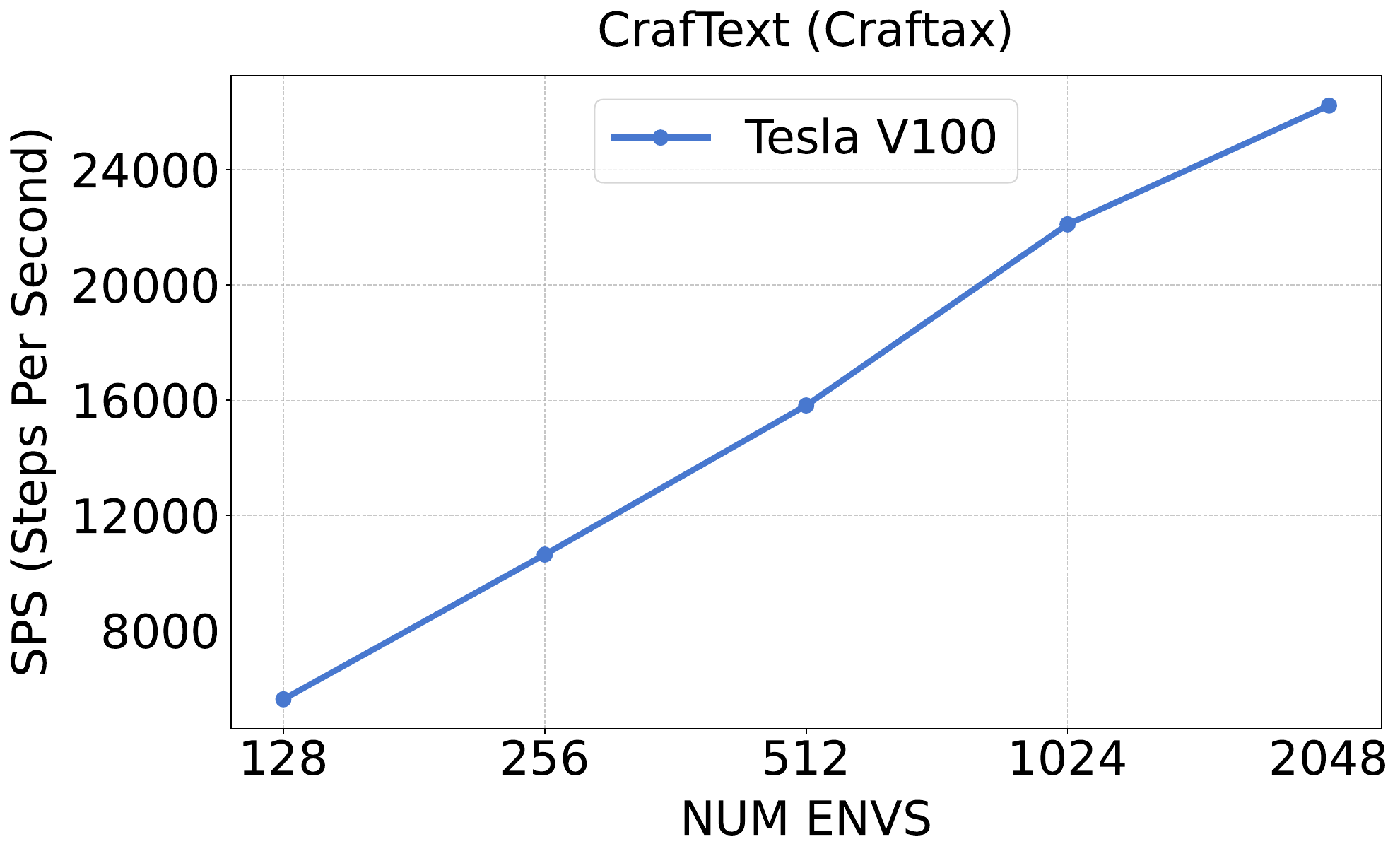}
    \caption{SPS of CrafText environment based on Craftax environment}
    \label{fig:sps}
\end{figure}

%------------------------------------------------------------------------------------------------
\section{Craftax} \label{app:craftax}
Craftax is a benchmark designed to facilitate research in open-ended RL. It builds upon the mechanics of the Crafter~\cite{hafner2021benchmarking} environment, providing a complex, procedurally generated world where agents must engage in deep exploration, long-term planning, and memory utilization while adapting to novel situations. The primary reward function in Craftax consists of achieving specific tasks that are grouped into four categories: 'Basic' (1 reward), 'Intermediate' (3 rewards), 'Advanced' (5 rewards), and 'Very Advanced' (8 rewards), enhancing the incentive structure for exploration and engagement. The observational input for agents is represented in two formats: a pixel-based input of size 110 × 130 × 3 and a symbolic representation with 512 dimension. The action space consists of 43 discrete actions, allowing agents to perform various interactions and movements within the environment. To succeed, agents must navigate diverse terrains, face 19 distinct creatures, and utilize various combat and crafting mechanics across nine unique procedurally generated floors.

\begin{figure}
    \centering
    \includegraphics[width=1.0\linewidth]{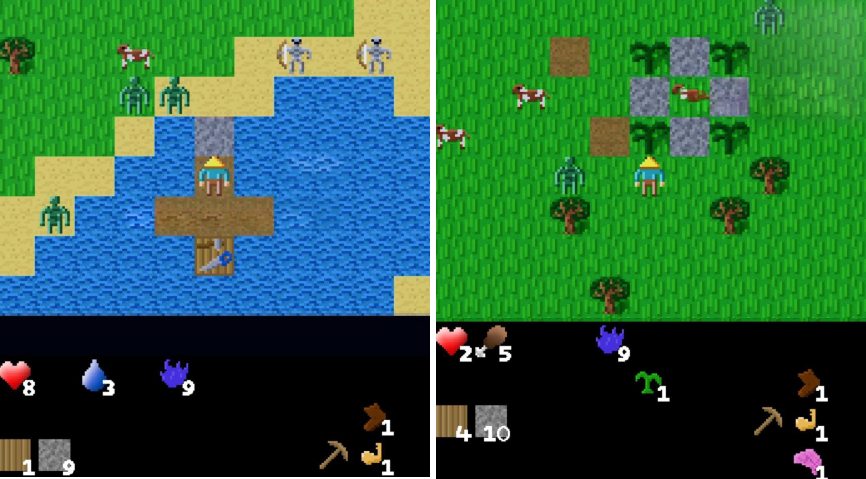}
    \caption{Examples of the visual observation input in Craftax, showcasing different perspectives of the environment provided to the agent for decision-making.}
    \label{fig:enter-label}
\end{figure}

\begin{figure*}[ht!]
    \centering
   % \hspace*{-1.5cm}  % Сдвигает изображение влево
    \includegraphics[width=0.9\textwidth]{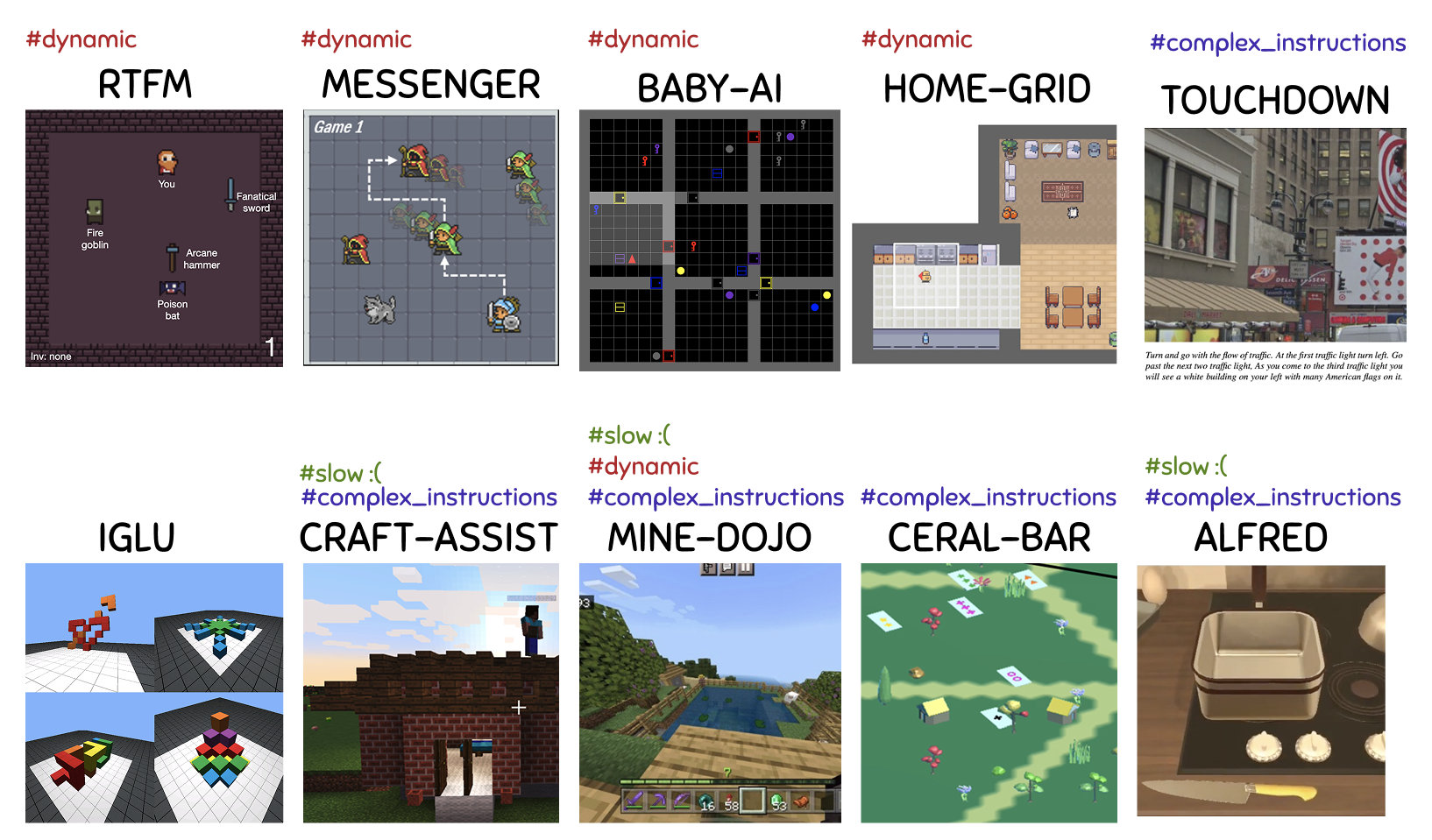}
    \caption{Visualizations of multimodal environments.}
    \label{fig:multi-environments}
\end{figure*}

\textbf{CrafText VS Craftax}.
CrafText builds upon Craftax by introducing instruction-based tasks, which require more than just a dataset of textual instructions. Each task is paired with a checker function that validates task execution at every environment timestep, providing continuous monitoring of the agent’s progress. This per-timestep validation is critical for ensuring correct task completion in dynamic environments. The checker signals the termination of an episode when a task is successfully solved and is tightly integrated with a task-based reward function, enabling immediate feedback essential for reinforcement learning.

To maintain the performance benefits of the original Craftax, CrafText implements these checker functions with XLA acceleration. This ensures that per-timestep validation and reward computation are computationally efficient, even in large-scale multimodal RL settings. Furthermore, by decoupling the {instruction-checker} dataset from the Craftax environment’s core implementation, CrafText supports seamless extensibility, enabling researchers to easily expand the benchmark with new tasks using the provided toolset.

CrafText is more than an environment—it is a comprehensive benchmark designed to evaluate natural language understanding and other capabilities in multimodal settings. Its evaluation protocol supports testing approaches using a hold-out dataset, which, while a correct methodology, remains underrepresented in the RL community. The protocol categorizes tasks into Conditional, Building, Localization, and Combination, each with varying difficulty levels (Easy, Medium, Hard). It also includes two state-of-the-art baselines, Dynalang and a large-scale PPO, both specifically adapted to the environment. The PPO baseline was trained for 1 billion environment steps in 12 hours on a single GPU, achieving over 20k steps per second (SPS), showcasing the benchmark’s scalability and accessibility to a broad range of researchers.

Finally, we emphasize that building upon prior research is central to scientific progress. Craftax itself was developed on top of Crafter and NetHack. Similarly, CrafText extends Craftax by introducing a benchmark for goal-driven natural language tasks in dynamic visual environments, supporting scalable task expansion and pioneering the use of XLA-accelerated multimodal benchmarks.

%------------------------------------------------------------------------------------------------
\section{Instruction Following Environments}
\label{app:env}

\begin{figure*}[ht!]
    \centering
    \begin{subfigure}[t]{0.48\textwidth}
    \includegraphics[width=1\textwidth]{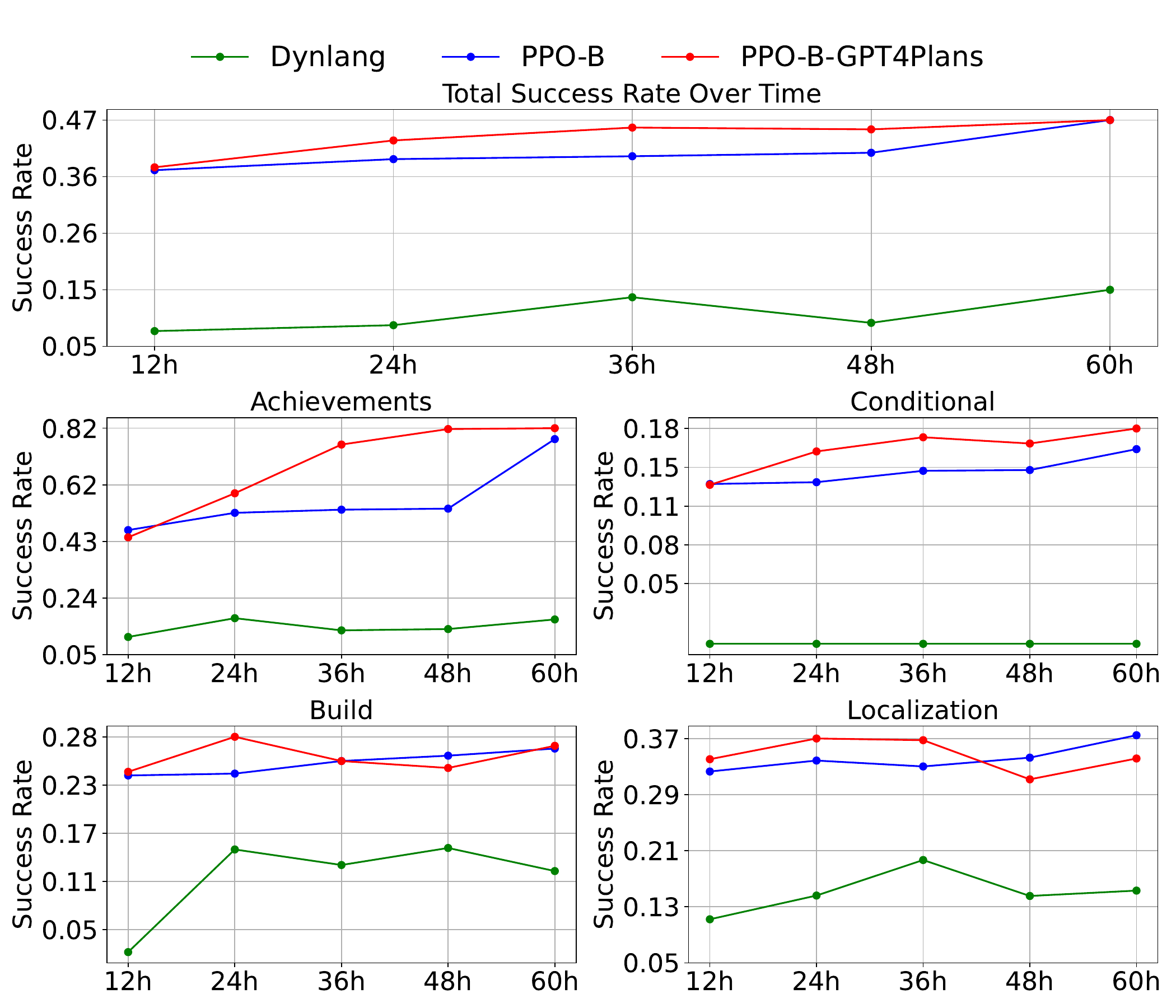} 
     \caption{Train}
    \end{subfigure}\hfill
    \begin{subfigure}[t]{0.48\textwidth}
    \includegraphics[width=1\textwidth]{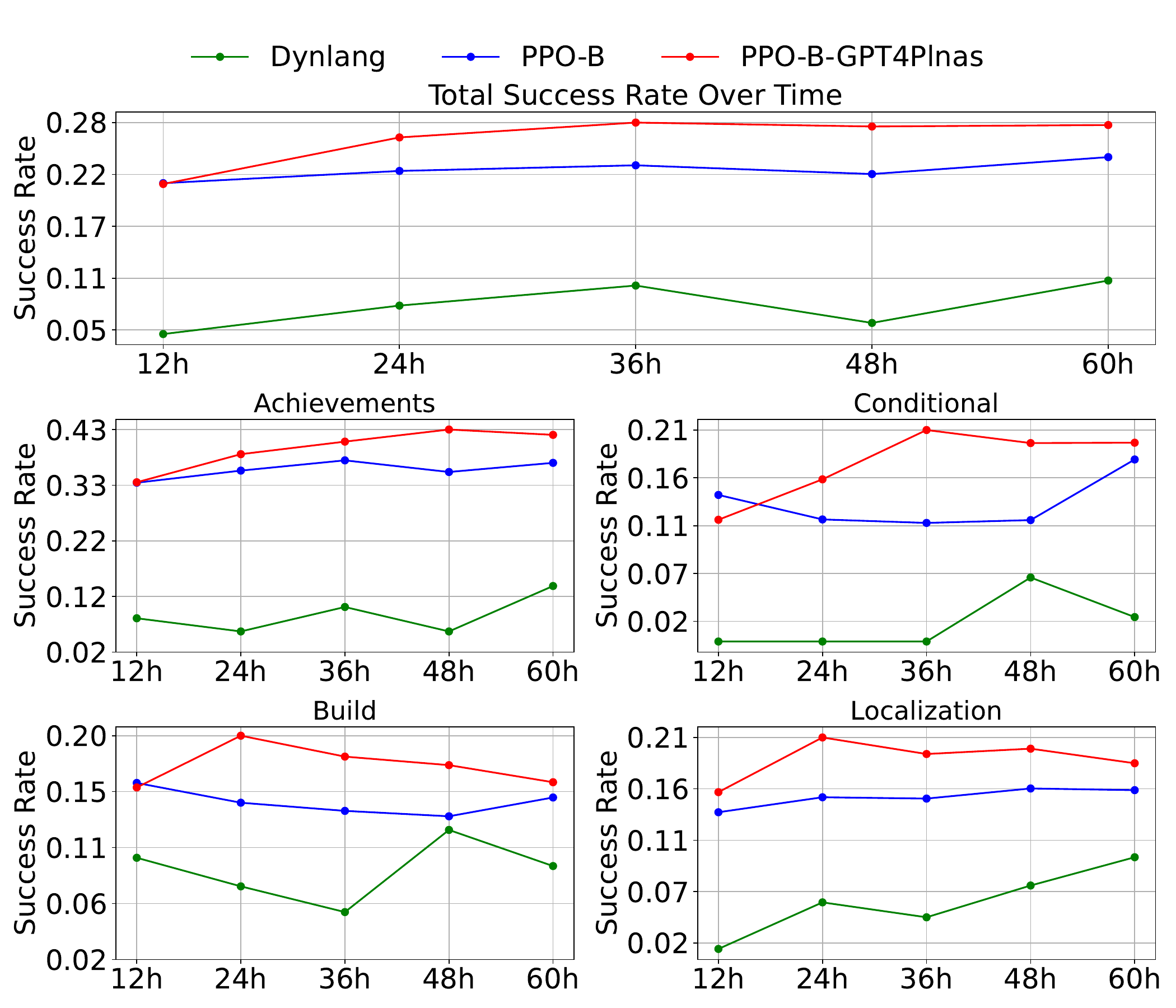}
     \caption{Other Params}
    \end{subfigure} \hfill
    
  \caption{The training and corresponding test (other parameters) curves for three baselines (PPO+BertEmb, PPO+BertEmb+GPT4Plans, and Dynalang)}
  \label{app:fig:three_pdfs}
\end{figure*}

\textbf{RTFM} is an environment where the agent must understand and apply procedural game instructions to navigate a dungeon. The agent reads procedurally generated manuals and uses the acquired knowledge to defeat enemies and interact with the game world. A key feature is the need for text comprehension in a dynamic setting.

\textbf{Messenger} presents a message-delivery challenge where the agent must navigate while avoiding obstacles and interacting with NPCs. The task requires choosing optimal routes and adapting to a dynamically changing environment, adding a layer of strategic complexity.

\textbf{Baby-AI} features a grid-based world where the agent follows simple language instructions. Tasks involve moving to specific locations and interacting with objects, requiring basic language comprehension and action planning in a discrete space.

\textbf{Home-Grid} simulates a household environment where the agent performs daily tasks. It moves through rooms, manipulates objects, and interacts with the surroundings based on given instructions. This environment models real-world scenarios with an emphasis on object management and action sequencing.

\textbf{Touchdown} is a city navigation simulation where the agent follows complex natural language instructions to traverse an urban environment. The setting consists of realistic 3D scenes, and successful task execution demands understanding spatial directions and adapting to detailed route descriptions.

\textbf{IGLU} focuses on spatial reasoning, requiring the agent to construct 3D structures by following verbal or textual instructions. The environment emphasizes precise execution of building tasks, making spatial understanding and object placement crucial.

\textbf{Craft-Assist} immerses the agent in a Minecraft-like world where it collaborates with humans to complete construction tasks. The agent interprets commands, interacts with the environment, and participates in cooperative work, posing challenges in language comprehension and behavior modeling.

\textbf{Mine-Dojo} provides an open-ended Minecraft world where the agent undertakes diverse tasks such as resource gathering, crafting, exploration, and combat. This environment requires complex planning and engagement with various mechanics, making it one of the most versatile interactive settings.

\textbf{CEREALBAR} is a collaborative instruction-following environment where a leader gives natural language instructions to a follower in a 3D game setting. The agent must accurately interpret and execute instructions to collect valid sets of cards. This setup emphasizes multi-agent coordination, real-time decision-making, and natural language understanding.

\textbf{ALFRED} places the agent in a photorealistic kitchen setting, where it executes household tasks such as cooking, cleaning, and organizing objects. The environment models complex object interactions and demands sequential execution of instructions while accounting for physical constraints.
%------------------------------------------------------------------------------------------------

\section{TRAINING: Curve Analysis} \label{app:curve}
We ran each baseline model—Dynalang, PPO+BertEmb (PPO-T), and PPO+BertEmb+GPT4Plan (PPO-T+) s—on the Easy portion of our dataset, training each model for 60 hours on an Tesla V100 GPU. Figure \ref{app:fig:three_pdfs} shows the result Success Rate, with checkpoints recorded every 12 hours on both the training and test sets. Overall, we observed similar performance with and without GPT-4 plans during training (overall 60h ±= 0.47). In contrast, Dynalang produced much weaker results (0.15), possibly because it requires additional training time due to its world model training process.

On the test set, the model incorporating GPT-4 plans achieved the best performance for every task type (overall 0.28, compared to PPO+BertEmb at 0.23 and Dynalang at 0.1). We attribute this improvement to the capability of large models like GPT-4 to simplify complex instructions. Moreover, on both the training and test sets, using GPT-4 plans yielded higher performance on conditional tasks (Training: 0.18 vs. 0.16 without plans; Test: 0.21 vs. 0.17 without plans). These results highlight that incorporating additional planning can help the agent solve puzzles and logical tasks that require understanding conditions expressed in text.

 %------------------------------------------------------------------------------------------------
\section{TRAINING: Prompt for GPT-4 plan generation}
\label{app:gpt4prompr}
Craftax is a virtual environment designed for exploration, crafting, and task completion. The procedurally generated world includes resources (trees, stones, coal, iron), interactive objects (crafting tables, furnaces, chests), and diverse terrains (water, grass, sand). The agent operates in this dynamic environment, performing actions such as moving, collecting resources, crafting, placing objects, and interacting with surroundings. Completing tasks often involves gathering resources, crafting items, and strategically placing objects.

The agent has a fixed set of discrete actions:
\begin{itemize}
    \item \textbf{Movement:} Navigate the map (up, down, left, right).
    \item \textbf{Resource Collection:} Chop trees, mine stones, or gather coal.
    \item \textbf{Crafting:} Create tools (e.g., pickaxes) or structures (e.g., furnaces).
    \item \textbf{Object Interaction:} Use objects (e.g., cook food in a furnace).
    \item \textbf{Placement:} Place crafted objects in specific locations.
\end{itemize}

\textbf{Task:} Create a step-by-step action plan (maximum 5 steps) for the agent in Craftax to achieve the following instruction:
\begin{quote}
``\{instruction\}''
\end{quote}

\textbf{Response Format:}
\begin{enumerate}
    \item Using only object names existing in the Craftax environment, provide the plan as a numbered list.
    \item Each step should outline a specific action or logical task for the agent, such as resource collection, crafting, or object placement.
    \item Keep steps clear, concise, and implementable in Craftax, with a maximum of 5 words per step.
\end{enumerate}

Check yourself!

\textbf{!ATTENTION!} Ensure the plan uses only object names and actions existing in Craftax. Replace any incorrect terms in the instruction with their correct Craftax equivalents.

%------------------------------------------------------------------------------------------------
\section{TRAINING: Analysis of Failure Cases} \label{app:fail_cases}

We analyze agent failures by evaluating the Success Rate for each instruction in our dataset and then aggregating these results based on key features: the types of objects the agent must interact with, specific task parameters (e.g., the length/width of items to be placed or the shape of the constructed structure), and the level of instruction complexity. By examining difficulties in interacting with various game elements, constructing complex shapes, and adapting to paraphrased instructions, we identify patterns of failure and determine where improvements can be made to enhance the agent’s task comprehension and execution.

\begin{figure}[ht!]
    \centering
    \includegraphics[width=0.97\linewidth]{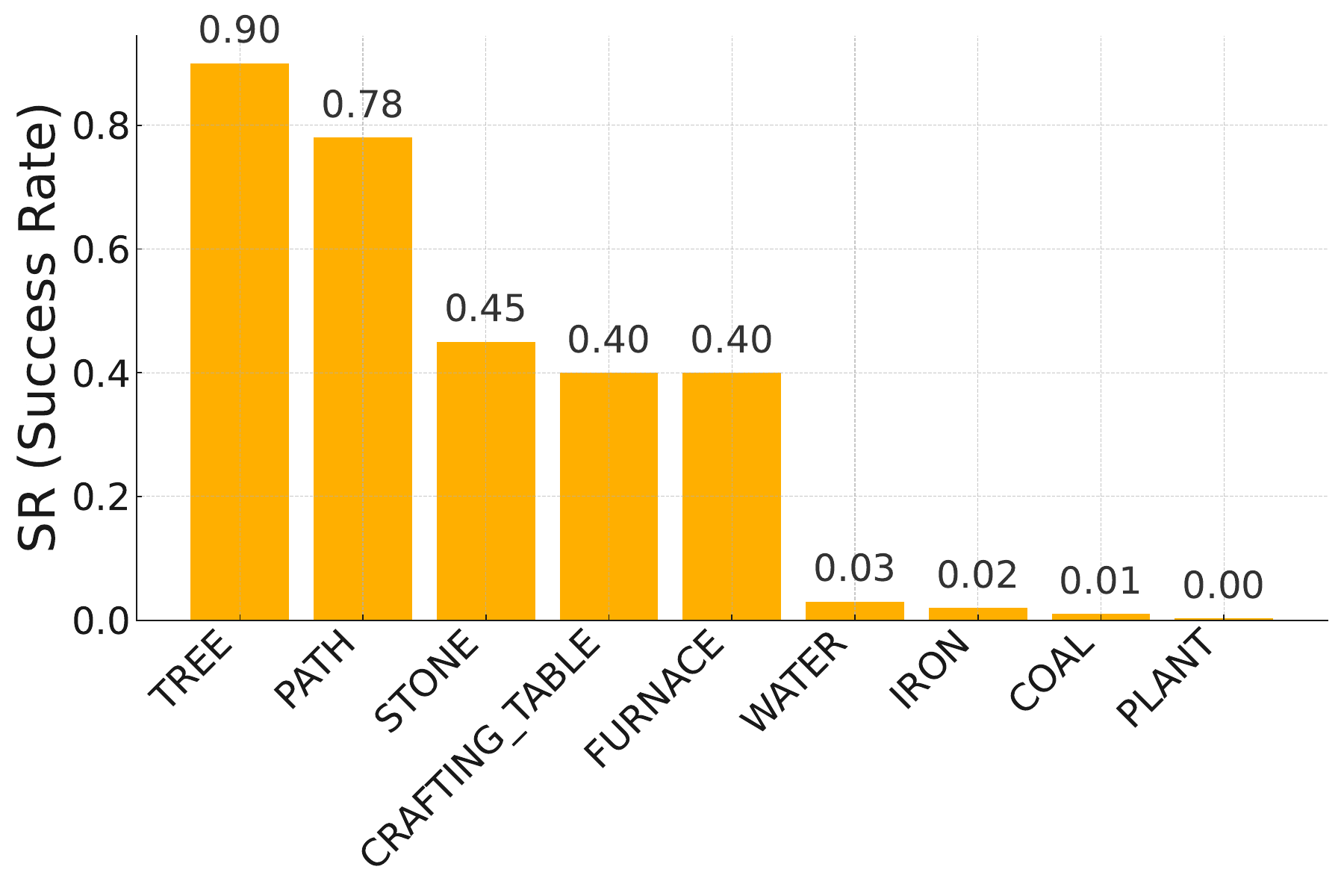}
    \caption{Aggregated SR for tasks involving different game elements, highlighting varying levels of interaction difficulty.}
    \label{fig:sr}
\end{figure}

The Figure~\ref{fig:sr} presents the aggregated SR for tasks involving different in-game elements. The highest SR is achieved for tasks requiring interaction with TREE (SR = 0.9), which is not surprising given that trees are abundant and do not demand specialized skills. In contrast, the lowest SR values occur for WATER (SR = 0.03), IRON (SR = 0.02), COAL (SR = 0.01), and PLANT (SR = 0.00). These more challenging tasks involve searching for specific objects and making use of appropriate inventory tools.

The agent struggles with building shapes larger than two blocks, making such configurations rare and rewards infrequent (see Tables~\ref{tab:line_sr} and \ref{tab:square-construct}). Although it can place two blocks in a line effectively (e.g., SR=0.78), performance drops sharply with three-block lines (SR=0.2) or diagonals (SR=0.07). Constructing squares is especially difficult (highest SR=0.09).

\begin{table}[ht!]
\centering
\small
\caption{Success rate averaged by goal (Line and Diagonal Construction)}
\label{tab:line_sr}
\begin{tabular}{@{}lccc@{}}
\toprule
\textbf{Block Type} & \textbf{Length} & \textbf{Diagonal} & \textbf{SR} \\
\midrule
STONE & 4 & False & 0.81 \\
CRAFTING\_TABLE & 2 & False & 0.78 \\
FURNACE & 2 & False & 0.72 \\
STONE & 2 & True & 0.67 \\
CRAFTING\_TABLE & 3 & False & 0.20 \\
CRAFTING\_TABLE & 3 & True & 0.07 \\
\bottomrule
\end{tabular}
\end{table}

\begin{table}[ht!]
\centering
\small
\caption{SR Averaged by Goal (Square Construction)}
\label{tab:square-construct}
\begin{tabular}{@{}lcc@{}}
\toprule
\textbf{Block Type} & \textbf{Side Size} & \textbf{Average\_SR} \\
\midrule
STONE & 2 & 0.09 \\
STONE & 3 & 0.004 \\
CRAFTING\_TABLE & 2 & 0.00 \\
FURNACE & 2 & 0.00 \\
\bottomrule
\end{tabular}
\label{tab:square_sr}
\end{table}

Finally, tasks using paraphrased instructions (e.g., synonyms instead of direct object names) present additional challenges for the agent. The average SR for tasks without paraphrases is 0.37, but it drops to 0.28 when paraphrased instructions are used. This underscores the difficulty of generalization in understanding varied instructions.

\subsection*{Greatest Challenges in the CrafText Environment}

\textbf{1. Environmental Features}

\textbf{Dynamic Instruction Following Task Setup.}  
An agent needs to interpret the same instruction in changing visual and spatial contexts.  
\textit{Experimental results (Table 2):} Despite extensive training, agent performance remains suboptimal even on the training set. This plateau highlights the difficulty of adapting to the dynamic and stochastic nature of the environment.  
Success rates (SR) on the training set: PPO-T = 0.4, PPO-T+ = 0.45, Dynalang = 0.15, FiLM = 0.43.

\textbf{Multi-Step Tasks with Implicit Preconditions.}  
Many tasks require multi-step action sequences where each subsequent step depends on successful completion of earlier ones. These may involve collecting resources, crafting intermediate items, and combining subgoals into a coherent plan.  
\textit{Experimental results (Appendix K):}  
(1) Tasks involving complex objects like IRON, COAL, and PLANT have very low SRs: 0.02, 0.01, and 0 respectively.  
(2) Even for simpler objects like the CRAFTING\_TABLE, increasing the line size from 2 to 3 blocks reduces SR from 0.78 to 0.2.

\vspace{1em}
\textbf{2. Linguistic Features}

\textbf{Linguistic Variation and Paraphrasing.}  
Each goal is expressed through multiple paraphrases that differ lexically and syntactically. The agent must understand diverse formulations, recognize equivalence, and map them to the same behavior.  
\textit{Experimental results (Table 2):}  
Performance drops significantly on paraphrased instructions:  
PPO-T: 0.4 $\rightarrow$ 0.36,  
PPO-T+: 0.45 $\rightarrow$ 0.35,  
Dynalang: 0.15 $\rightarrow$ 0.05,  
FiLM: 0.43 $\rightarrow$ 0.35.

\textbf{Examples: Goal = Make a line with size 2 from Crafting Table}

\begin{center}
\begin{tabular}{>{\raggedright\arraybackslash}p{6cm}|c}
\hline
Instruction & SR \\
\hline
Make a line of 2 blocks using table & 0.85 \\
Construct a row of 2 pieces with the crafting station & 0.46 \\
Arrange a sequence of 2 blocks with the crafting platform & 0.33 \\
\hline
\end{tabular}
\end{center}

\vspace{1em}
\textbf{3. Generalization to Novel Combinations}

\textbf{Test New Objects Split.}  
Includes instructions with new combinations of objects, spatial relations, and parameters. The agent must generalize beyond memorized templates and recombine known elements in novel ways.  
\textit{Experimental results (Table 2):}  
All baselines show strong performance drops:  
PPO-T: 0.4 $\rightarrow$ 0.22,  
PPO-T+: 0.45 $\rightarrow$ 0.28,  
Dynalang: 0.15 $\rightarrow$ 0.10,  
FiLM: 0.43 $\rightarrow$ 0.26.

%------------------------------------------------------------------------------------------------
\section{TRAINING: Hyperparameters}\label{app:ppo_params}

\begin{table*}[h!]
\centering
\begin{tabular}{lccccc}
\hline
\textbf{Model} & \textbf{Trained} & \textbf{Achievements (one)} & \textbf{Achievements (combo)} & \textbf{New objects} \\
\hline
DeepSeek & X & 0.15 & 0.07 & 0.06 \\
MISTRAL & X & \textbf{0.21} & \textbf{0.10} & \textbf{0.09} \\
Qwen & X & \textbf{0.21} & \textbf{0.10} & 0.07 \\
PPO-T & YES & 0.78 & 0.55 & 0.34 \\
PPO-T+ & YES & 0.87 & 0.70 & \textbf{0.43} \\
Dynalang & YES & -- & 0.17 & 0.14 \\
FiLM & YES & 0.85 & 0.76 & 0.38 \\
\hline
\end{tabular}
\caption{Comparison of models on the EASY part of the CrafText dataset: Achievements (one) involve solving single atomic tasks, Achievements (combo) require completing multiple achievements together, and New objects measure performance on unseen combinations of achievements.}
\label{tab:easy_exp}
\end{table*}

The training of our models relies on carefully selected hyperparameters for both the language processing module and the reinforcement learning  agent. The Dynalang model utilizes a deep language MLP with 5 layers of 1024 units each, a GRU with 4096 recurrent units, and a total parameter count of 281 million, optimized using Adam with a learning rate of $1\times10^{-4}$ and SiLU activations.

For the RL component in PPO-T, PPO-T+, and FiLM, we employ PPO with 1024 parallel environments and a training horizon of 1 billion timesteps. PPO training uses a learning rate of 0.0002, a discount factor of 0.99, GAE lambda of 0.8, and 4 update epochs per batch, with 8 minibatches per update. We apply clipping with $\epsilon=0.2$, an entropy coefficient of 0.01 to encourage exploration, and a value function coefficient of 0.5 for stabilizing value estimates. The agent network uses \texttt{tanh} activations with a hidden layer size of 512 units. Additionally, we enable optimistic resets with a reset ratio of 16 to improve exploration efficiency. Instructional inputs are processed using a DistilBERT encoder, with right-padding and truncation applied during tokenization.

\begin{table}[ht!]
\centering 
\small
\caption{Dynalang Training Hyperparameters}
\begin{tabular}{ll}
\hline
\textbf{Parameter Name}          & \textbf{Default Value}          \\ \hline
Language MLP Layers              & 5                               \\ %\hline
Language MLP Units               & 1024                            \\ %\hline
Batch Size                       & 16                              \\ %\hline
Batch Length                     & 256                             \\ %\hline
Train Ratio                      & 32                              \\ %\hline
GRU Recurrent Units              & 4096                            \\ %\hline
Total Model Parameters           & 281M                            \\ %\hline
Optimizer                        & Adam                            \\
Learning Rate                    & 1e-4                            \\
Epsilon                          & 1e-8                            \\
Activation                       & SiLU                            \\
\hline
\end{tabular}
\label{tab:model_hyperparameters}
\end{table}

\begin{table}[ht!]
\centering 
\small
\caption{PPO Training Hyperparameters}
\begin{tabular}{ll}
\hline
\textbf{Parameter Name}          & \textbf{Default Value}          \\ \hline
Number of Environments           & 1024                             \\ %\hline
Total Training Timesteps         & 1 000,000,000                     \\% \hline
Learning Rate                    & 0.0002                          \\ %\hline
Steps per Environment            & 100                             \\ %\hline
Update Epochs                    & 4                               \\ %\hline
Minibatches per Update           & 8                               \\ %\hline
Discount Factor                  & 0.99                            \\ %\hline
GAE Lambda                       & 0.8                             \\ %\hline
Clipping Epsilon                 & 0.2                             \\ %\hline
Entropy Coefficient              & 0.01                            \\ %\hline
Value Function Coefficient       & 0.5                             \\ %\hline
Max Gradient Norm                & 1.0                             \\ %\hline
Activation Function              & tanh                            \\ %\hline
Anneal Learning Rate             & True                            \\ %\hline
Layer Size                       & 512                             \\ %\hline
Optimistic Resets Enabled        & True                            \\ %\hline
Optimistic Reset Ratio           & 16                              \\ %\hline
Exploration Epochs               & 4                               \\ %\hline
Instruction Encoder Model        & DistilBERT                      \\
Encoder Tokenizer Padding        & right                           \\
Encoder Tokenizer Truncation     & right                           \\
\hline
\end{tabular}
\label{tab:ppo_hyperparameters}
\end{table}

\section{TRAINING: EASY-dataset experiments}\label{app:ppo_params}

We introduce three zero-shot baselines for our EASY set, using popular LLMs to predict actions based on instructions and structured descriptions of the current agent observation, including objects and mobs coordinates. The models used are Qwen/QWQ-32B (Qwen), DeepSeek-R1-Distill-Qwen-32B (DeepSeek), and Mistral-Small-24B-Instruct-2501 (MISTRAL).

We evaluate them alongside trainable baselines (PPO-T, PPO-T+, Dynalang, FiLM) on two categories within the EASY set: (1) \textit{Achievement (one)} — single-step goals (e.g., ``Make a crafting table''); (2) \textit{Achievement (combo)} — multi-step goals involving several achievements (e.g., ``Make a crafting table and craft a stone sword'').

We also report results on \textit{New objects}, which contains instructions requiring the agent to complete unseen combinations of achievements within a single episode.

\textbf{Zero-shot LLMs show basic environment understanding but struggle under dynamic conditions.}

In the zero-shot setting, language models achieve a score of \( 0.21 \) on \textit{Achievements (one)}, suggesting they capture some aspects of the environment’s mechanics and object relationships. Their performance on \textit{Achievements (combo)} is lower (\( 0.10 \)), indicating limited ability to handle composite instructions requiring long-horizon decision-making. Overall performance remains low, emphasizing the limitations of LLMs in decision-making under dynamic and partially observable conditions.

\textbf{Generalization on unseen tasks.}

The evaluation on \textit{New objects} is the only part of the dataset where it is valid to compare trained agents and language models, as both encounter these combinations for the first time. Here, trained agents such as PPO-T+ (\( 0.43 \)) and FiLM (\( 0.38 \)) significantly outperform zero-shot models, whose scores remain in the range \( 0.06\text{--}0.09 \). At the same time, we observe that trained RL models experience a notable drop in performance compared to the training set — almost a twofold decrease. In contrast, LLMs, as expected, exhibit only minor changes in performance.

Additionally, the performance of the evaluated LLMs could be significantly enhanced with better prompt tuning and a richer representation of the agent’s input using environment state. Prior work, such as Voyager~\cite{wang2023voyager}, has demonstrated how structured prompting can improve LLM-based agents in game-like environments. However, our study does not aim to optimize LLM prompting for this domain but rather to establish baseline zero-shot performance.

\end{document}